\documentclass{article}

\usepackage{microtype}
\usepackage{graphicx}
\usepackage{subcaption}

\usepackage{booktabs} 
\usepackage{arydshln}
\usepackage{ulem}
\usepackage{float}
\usepackage{multirow}
\usepackage{amsmath, amsfonts, amssymb}
\usepackage{hyperref}

\usepackage{xcolor}

\definecolor{ForestGreen}{rgb}{0.0, 0.5, 0.0}
\definecolor{NavyBlue}{rgb}{0, 0.44, 0.75}

\usepackage[accepted]{icml2025}

\usepackage{amsmath}
\usepackage{amssymb}
\usepackage{mathtools}
\usepackage{amsthm}
\usepackage{xspace}

\usepackage{xcolor}
\definecolor{Graylight}{gray}{0.9}
\definecolor{Gray}{gray}{1.0}
\usepackage{colortbl}
\usepackage{multirow}
\usepackage{makecell}
\usepackage{bm}

\usepackage{pifont}
\definecolor{mygreen}{RGB}{9,136,66}
\newcommand{\cmark}{\textcolor{mygreen}{\ding{51}}}%
\newcommand{\xmark}{\textcolor{red!50}{\ding{55}}}%

\usepackage[capitalize,noabbrev]{cleveref}

\newcommand{\methodname}{VideoRoPE\xspace}

\theoremstyle{plain}

\theoremstyle{definition}

\theoremstyle{remark}

\usepackage[textsize=tiny]{todonotes}

\icmltitlerunning{\methodname: What Makes for Good Video Rotary Position Embedding?}

\begin{document}

\twocolumn[
\icmltitle{\methodname: What Makes for Good Video Rotary Position Embedding?}



\icmlsetsymbol{equal}{*}

\begin{icmlauthorlist}
\icmlauthor{Xilin Wei}{equal,fdu,ailab}
\icmlauthor{Xiaoran liu}{equal,fdu,ailab,inno}
\icmlauthor{Yuhang Zang}{ailab}
\icmlauthor{Xiaoyi Dong}{ailab,cuhk}
\icmlauthor{Pan Zhang}{ailab}
\icmlauthor{Yuhang Cao}{ailab}
\icmlauthor{Jian Tong}{ailab}
\icmlauthor{Haodong Duan}{ailab}
\icmlauthor{Qipeng Guo}{ailab}
\icmlauthor{Jiaqi Wang}{ailab}
\icmlauthor{Xipeng Qiu}{fdu,ailab,inno}
\icmlauthor{Dahua Lin}{ailab,cuhk,cpii}
\end{icmlauthorlist}

\icmlaffiliation{fdu}{Fudan University, Shanghai, China}
\icmlaffiliation{ailab}{Shanghai AI Laboratory, Shanghai, China}
\icmlaffiliation{inno}{Shanghai Innovation Institute, Shanghai, China}
\icmlaffiliation{cuhk}{The Chinese University of Hong Kong}
\icmlaffiliation{cpii}{CPII under InnoHK}


\icmlcorrespondingauthor{Yuhang Zang}{zangyuhang@pjlab.org.cn}
\icmlcorrespondingauthor{Qipeng Guo}{guoqipeng@pjlab.org.cn}
\icmlcorrespondingauthor{Jiaqi Wang}{wangjiaqi@pjlab.org.cn}


\vskip 0.3in
]



\printAffiliationsAndNotice{\icmlEqualContribution} 

\begin{abstract}

While Rotary Position Embedding (RoPE) and its variants are widely adopted for their long-context capabilities, the extension of the 1D RoPE to video, with its complex spatio-temporal structure, remains an open challenge.
This work first introduces a comprehensive analysis that identifies four key characteristics essential for the effective adaptation of RoPE to video, which have not been fully considered in prior work.
As part of our analysis, we introduce a challenging V-NIAH-D (Visual Needle-In-A-Haystack with Distractors) task, which adds periodic distractors into V-NIAH.
The V-NIAH-D task demonstrates that previous RoPE variants, lacking appropriate temporal dimension allocation, are easily misled by distractors.
Based on our analysis, we introduce \textbf{VideoRoPE}, with a \textit{3D structure} designed to preserve spatio-temporal relationships.
VideoRoPE features \textit{low-frequency temporal allocation} to mitigate periodic oscillations, a \textit{diagonal layout} to maintain spatial symmetry, and \textit{adjustable temporal spacing} to decouple temporal and spatial indexing.
VideoRoPE consistently surpasses previous RoPE variants, across diverse downstream tasks such as long video retrieval, video understanding, and video hallucination. Our code is available at \href{https://github.com/Wiselnn570/VideoRoPE}{https://github.com/Wiselnn570/VideoRoPE}.


\end{abstract}
\vspace{-6mm}
\section{Introduction}

Rotary Position Embedding (RoPE) \cite{su2024roformer} helps Transformer models understand word order by assigning each token a unique positional `marker' calculated using a mathematical rotation matrix.
RoPE has advantages in long-context understanding \cite{ding2024longrope}, and continues to be a default choice in leading Large Language Models (LLMs) like the LLaMA \cite{touvron2023llamaopenefficientfoundation,touvron2023llama,dubey2024llama} and QWen \cite{yang2024qwen2,yang2024qwen25} series.

The original RoPE implementation (Vanilla RoPE) \cite{su2024roformer} is designed for sequential 1D data like text. However, recent Video Large Language Models (Video LLMs) \cite{2023videochat,lin2023video,chen2024sharegpt4video,maaz2024videochatgptdetailedvideounderstanding,zhang2024longva,wang2024longllavascalingmultimodalllms,chen2024longvilascalinglongcontextvisual,internlmxcomposer2_5_OL} process video, which has a more complex spatio and temporal structure.
As shown in Tab. \ref{tab:pe_compare}, although several RoPE-based approaches \cite{gao2024tc,wang2024qwen2} have been proposed to support video inputs, these variants exhibit limitations and do not fully satisfy the following key characteristics:

\begin{table}[t]
\tiny
\centering
\caption{Comparison between different RoPE variants for Video Large Language Models (Video LLMs).}
\vspace{-2pt}
\tabcolsep=0.1cm
\begin{tabular}{lcccc}
\toprule
 & \makecell[c]{\textbf{2D/3D} \\ \textbf{Structure}} & \makecell[c]{\textbf{Frequency} \\ \textbf{Allocation}} & \makecell[c]{\textbf{Spatial} \\ \textbf{Symmetry}} & \makecell[c]{\textbf{Temporal} \\ \textbf{Index Scaling}} \\
\midrule
\makecell[l]{Vanilla RoPE \cite{su2024roformer}} & \xmark & \xmark & \xmark & \xmark \\
\makecell[l]{TAD-RoPE \cite{gao2024tc}} & \xmark & \xmark & \xmark & \cmark \\
\makecell[l]{RoPE-Tie \cite{kexuefm10040}} & \cmark & \xmark & \cmark & \xmark \\
\makecell[l]{M-RoPE \cite{wang2024qwen2}} & \cmark & \xmark & \xmark & \xmark \\
\midrule
\rowcolor[HTML]{F2F3F5}
\methodname (Ours) & \cmark & \cmark & \cmark & \cmark \\
\bottomrule
\end{tabular}
\label{tab:pe_compare}
\vspace{-12pt}
\end{table}

\begin{figure}
\centering
\includegraphics[width=0.88\linewidth]{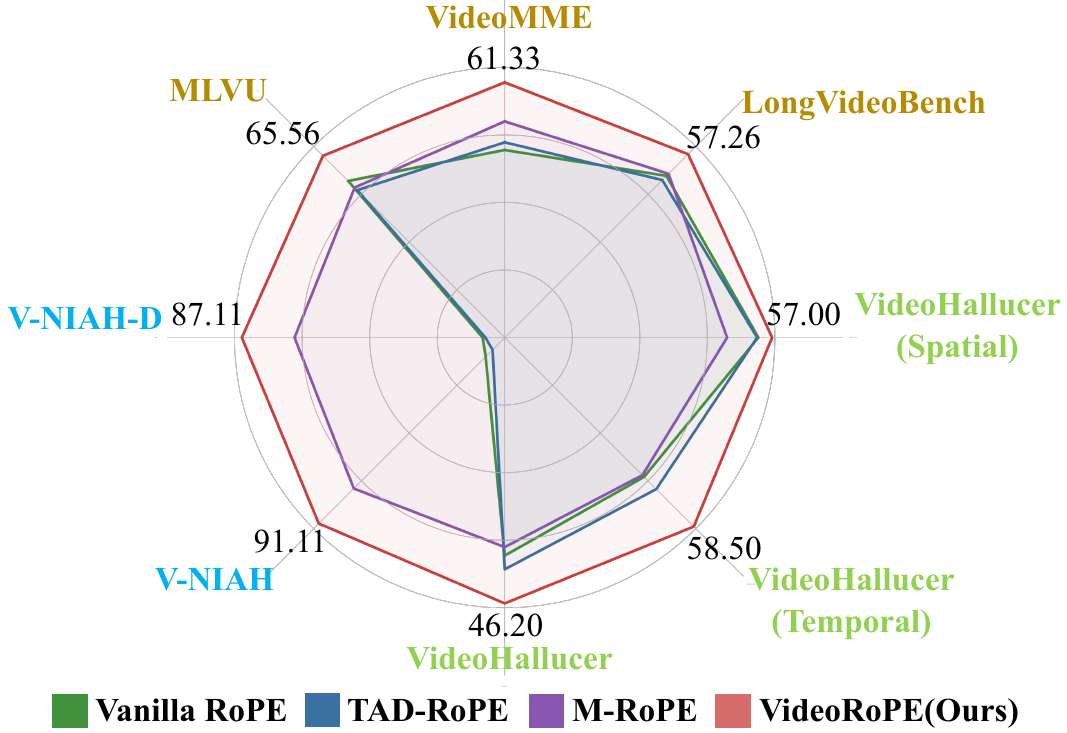}
\vspace{-6pt}
\caption{\methodname outperforms RoPE variants on benchmarks.}
\label{fig:radar}
\vspace{-12pt}
\end{figure}

\begin{figure*}[ht]
    \centering
    \includegraphics[width=0.96\linewidth]{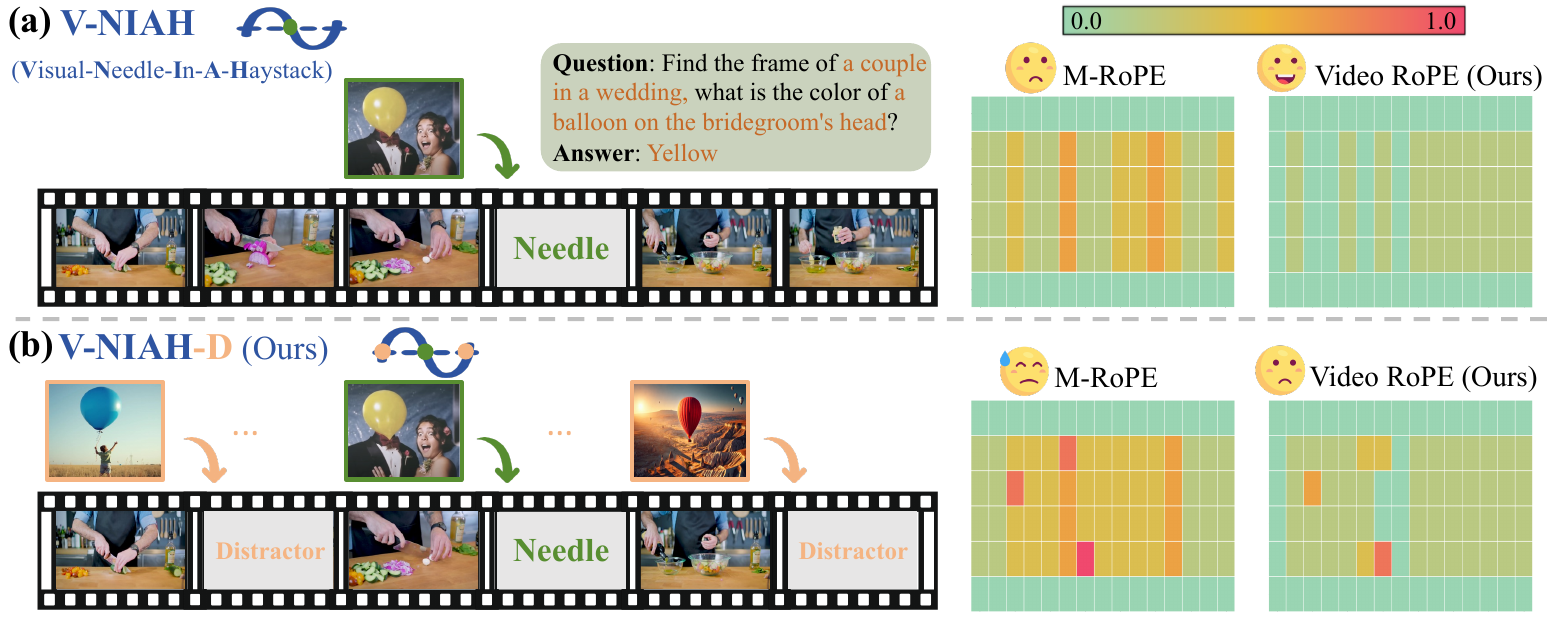}
    \vspace{-6pt}
    \caption{\footnotesize \textbf{Left:} To demonstrate the importance of frequential allocation, based on VIAH (\textbf{a}) we present a more challenging V-NIAH-D task (\textbf{b}) that similar images are inserted as distractors.
    \textbf{Right:} Compared to M-RoPE, our \methodname is more robust in retrieval and is less affected by distractors. See Fig. \ref{fig:v-niah-and-d} in the Experiments section for details on the horizontal and vertical axes.
    }
    \label{fig:v-ruler}
    \vspace{-12pt}
\end{figure*}

\textbf{(1) 2D/3D Structure.} Some existing Video LLMs direct flatten the video frame into 1D embeddings and apply the 1D structure RoPE \cite{su2024roformer,gao2024tc}.
These solutions fail to capture video data's inherent 2D or 3D (temporal ($t$), horizontal ($x$), and vertical ($y$)) structure, thus hindering explicit spatial and temporal representation.


\textbf{(2) Frequency Allocation.} Previous approaches such as M-RoPE used in QWen2-VL \cite{wang2024qwen2} employ 3D structure, dividing feature dimensions into distinct subsets for ($t$, $x$, $y$) encoding, respectively.
How to determine the optimal allocation of these dimension subsets, and their associated frequencies 
\footnote{In RoPE, frequencies are determined by $\beta^{-2n/d}$, where $\beta$ is a constant, $n$ is the dimension index, $d$ is the total number of dimensions. Thus, choosing which dimensions represent $t$, $x$, and $y$ directly determines the frequencies used for each.} are not well studied.
Some previous work allocates the lower dimensions corresponding to the high frequency to represent the $t$.
However, the temporal dimension $t$ is significantly tortured by periodic oscillation, and distant positions may have the same embeddings.

We present a simple setting to verify this point.
Based on the previous long-video retrieval task V-NIAH (Visual Needle-In-A-Haystack) \cite{zhang2024longva}, we insert several similar images that do not affect the question's answer before and after the needle image as distractor \cite{hsieh2024ruler,yuan2024lv}, forming a new task, V-NIAH-D (Visual Needle-In-A-Haystack with Distractors).
As shown in Fig. \ref{fig:v-ruler}, we find that previous M-RoPE is misled by distractors, showing a significant performance decline from V-NIAH to V-NIAH-D.
Our observation demonstrates that the periodic oscillation reduces Video LLMs' robustness.

\textbf{(3) Spatial Symmetry.} The distance between the end of the precedent textual input and the start of visual input equals the distance between the end of visual input and the start of subsequent textual input~\cite{kexuefm10352}. Such a symmetry ensures that the visual input receives equal contextual influence from both the preceding and subsequent textual information.

\textbf{(4) Temporal Index Scaling.} Spatial and temporal dimensions often exhibit different granularities (e.g., a unit change in $x$/$y$ differs from a unit change in $t$) \cite{gao2024tc}.
Employing varying index intervals in positional encoding allows for dimension-specific encoding, capturing diverse scales and enhancing efficiency.

Driven by our analysis, we present a new video position embedding strategy, \textbf{\methodname}, which can simultaneously satisfy the four properties in Tab. \ref{tab:pe_compare}.
Specifically, we use a 3D structure to model spatiotemporal information, allocating higher dimensions (lower frequencies), to the temporal axis (\textbf{L}ow-frequency \textbf{T}emporal \textbf{A}llocation, \textbf{LTA}) to prioritize temporal modeling.
The right panel of Fig. \ref{fig:v-ruler} demonstrates that our LTA allocation mitigates oscillations and exhibits robustness to distractors in the V-NIAH-D task.
We further employ a \textbf{D}iagonal \textbf{L}ayout (\textbf{DL}) design to ensure spatial symmetry and preserve the relative positioning between visual and text tokens.
Regarding temporal index scaling, we propose \textbf{A}djustable \textbf{T}emporal \textbf{S}pacing (\textbf{ATS}), where a hyper-parameter controls the relative temporal spacing of adjacent visual tokens.
In summary, our proposed position encoding scheme demonstrates favorable characteristics for modeling video data, yielding a robust and effective representation of positional information.

Overall, the contributions of this work are summarized as:

\textbf{(1)} We present an analysis of four key properties essential for RoPE when applied to video. Motivated by this analysis, we propose \methodname including Low-frequency Temporal Allocation (LTA), Diagonal Layout (DL), and Adjustable Temporal Spacing (ATS) to satisfy all four properties.

\textbf{(2)} We introduce the challenging V-NIAH-D task to expose the drawbacks of current position embedding designs regarding frequency allocation. We reveal that existing Video LLMs are easily misled to frequency-based distractors.

\textbf{(3)} Extensive experiments demonstrate that \methodname consistently achieves superior performance compared to other RoPE variants. For example, \methodname outperforms previous M-RoPE on long video retrieval (\textbf{+12.4} on V-NIAH, \textbf{+12.4} on V-NIAH-D), video understanding (\textbf{+2.9} on LongVideoBench, \textbf{+4.5} on MLVU, \textbf{+1.7} on Video-MME) and hallucination (\textbf{+11.9} on VideoHallucer) benchmarks.
\vspace{-3mm}
\section{Related Work}

\noindent \textbf{RoPE (Rotary Position Embedding).}
RoPE~\cite{su2024roformer} is a pivotal mechanism for encoding positional information in LLM long-context modeling. Using a rotation matrix, RoPE unifies the advantages of both absolute and relative positional embedding schemes. 
In RoPE design, different feature dimensions are embedded with position information based on Trigonometric functions $\sin$ and $\cos$ with different frequencies~\cite{peng2023yarn,liu2023scaling}.
Lower dimensions correspond to higher frequency given larger values of base frequency.
The simplicity and effectiveness of RoPE have led to its widespread adoption in leading LLMs~\cite{touvron2023llamaopenefficientfoundation, yang2024qwen2, gemmateam2024gemmaopenmodelsbased,cai2024internlm2, Sun2024MOSS}.

\noindent \textbf{Extending RoPE to Multi-Modal Data.}
Extending RoPE to multi-modal or Video LLMs typically follows two approaches.
One approach directly applies standard RoPE, flattening visual tokens and treating text and visual tokens as a single 1D sequence. Although variants (e.g., TAD-RoPE \cite{gao2024tc}) introduce enhancements in indexing and attention mechanisms, these 1D RoPE variants overlook the spatiotemporal structure of video and inherent inter-modal differences \cite{kexuefm10040,kexuefm10352,wang2024qwen2}.
In contrast, several studies have explored incorporating structural information to formulate the 2D/3D RoPE.
For example, some previous works \cite{agrawal2024pixtral12b,wang2024qwen2} integrate RoPE-2D into visual encoders to improve spatial representation, particularly for resolution scaling.
Based on the RoPE-Tie \cite{kexuefm10040}, M-RoPE \cite{wang2024qwen2} used in QWen2-VL further generalizes RoPE to three dimensions to model both temporal and spatial dynamics.
While effective, M-RoPE exhibits limitations, such as struggles with distractors in our V-NIAH-D task.
This work presents a comprehensive analysis of the important characteristics essential for extending RoPE to video and proposes \methodname according to our analysis.

\section{Analysis}

\textbf{3D Structure.}
The vanilla RoPE defines a matrix $\bm{A}_{t_1,t_2}$ that represents the relative positional encoding between two positions $t_1$ and $t_2$ in a 1D sequence:
\begin{equation}\label{eq:vanilla_rope}
\begin{aligned}
\bm{A}_{t_1,t_2}&=\left(\bm{q}_{t_1}\bm{R}_{t_1}\right){\left(\bm{k}_{t_2}\bm{R}_{t_2}\right)}^\top
= \bm{q}_{t_1}\bm{R}_{\Delta t}\bm{k}_{t_2}^\top,
\end{aligned}
\end{equation}
where $\Delta t=t_1-t_2$, the symbols $\bm{q}_{t_1}$ and $\bm{k}_{t_2}$ are the query and key vectors at positions $t_1$ and $t_2$.
The \textit{relative rotation matrix} $\bm{R}_{\Delta t}$ is defined as $\bm{R}_{\Delta t} = \exp(\Delta ti\theta_{n})$, while $i$ is the imaginary unit, $\theta_{n} = \beta^{-2n/d}$ is the frequency of rotation applied to a specific $n$-th pair of $d$ dimensions ($n=0,\ldots,d/2-1$), and $\beta$ is the frequency base parameter.
The vanilla RoPE uses $d=128$, thus $n=0,\ldots,63$.
Consequently, the $\bm{A}_{t_1,t_2}$ in Eq. (\ref{eq:vanilla_rope}) can be extended as:
\begin{equation}\label{equ:rope}
\vspace{-6pt}
\resizebox{0.5\textwidth}{!}{$
\scriptsize
\left(
\begin{array}{c}
q^{(0)}\\q^{(1)}\\\vdots\\q^{(126)}\\q^{(127)}
\end{array}
\right)^{\top}
\left(
\begin{array}{ccccc}
\cos{\theta_0\Delta t} & -\sin{\theta_0\Delta t} & \cdots & 0 & 0 \\ 
\sin{\theta_0\Delta t} & \cos{\theta_0\Delta t} & \cdots & 0 & 0 \\ 
\vdots & \vdots & \ddots & \vdots & \vdots \\  
0 & 0 & \cdots & \cos{\theta_{63}\Delta t} &  \sin{\theta_{63}\Delta t} \\  
0 & 0 & \cdots & \sin{\theta_{63}\Delta t} & \cos{\theta_{63}\Delta t} 
\end{array}
\right)
\left(
\begin{array}{c}
k^{(0)}\\k^{(1)}\\\vdots\\k^{(126)}\\k^{(127)}
\end{array}
\right)
$}
\end{equation}

While the vanilla RoPE operates on 1D sequences, it can also be applied to higher-dimensional input by flattening the input into a 1-D sequence.
However, the flattening process discards crucial neighborhood information, increases the sequence length, and hinders the capture of long-range dependencies.
Therefore, preserving the inherent 3D structure is essential when adapting RoPE for video data.
Some recent RoPE-variants (e.g., M-RoPE in Qwen2-VL \cite{wang2024qwen2}) incorporate the $3$D structure.
The corresponding relative matrix $\bm{A}_{(t_1,x_1,y_1)}$ is computed as:
\begin{equation}
\bm{A}_{(t_1,x_1,y_1),(t_2,x_2,y_2)}=\bm{q}_{(t_1,x_1,y_1)}\bm{R}_{\Delta t,\Delta x,\Delta y}\bm{k}_{(t_2,x_2,y_2)}^\top,
\end{equation}
where $\Delta t=t_1-t_2$, $\Delta x=x_1-x_2$, and $\Delta y=y_1-y_2$.
M-RoPE divides the $d=128$ feature dimensions into 3 groups: the first 32 for temporal positions ($t$), the middle 48 for horizontal positions ($x$), and the last 48 for vertical positions ($y$). As shown in Eq~(\ref{equ:mrope}), $\bm{A}_{(t_1,x_1,y_1),(t_2,x_2,y_2)}$ in M-RoPE is extended as:
\begin{equation}
\vspace{-6pt}
\resizebox{0.5\textwidth}{!}{$
\scriptsize
\begin{gathered}
\underbrace{\begingroup
\setlength\arraycolsep{1pt}
\begin{pmatrix}q^{(0)}\\q^{(1)}\\q^{(2)}\\q^{(3)}\\\vdots\\q^{(30)}\\q^{(31)}\end{pmatrix}^\top
\begin{pmatrix}
\cos{\theta_0\Delta t}& -\sin{\theta_0\Delta t}&0&0&\cdots&0&0\\
\sin{\theta_0\Delta t}&\cos{\theta_0\Delta t}&0&0&\cdots&0&0 \\
0&0&\cos{\theta_1\Delta t}& -\sin{\theta_1\Delta t}&\cdots&0&0\\
0&0&\sin{\theta_1\Delta t}&\cos{\theta_1\Delta t}&\cdots&0&0 \\ 
\vdots&\vdots&\vdots&\vdots&\ddots&\vdots&\vdots\\
0&0&0&0&\cdots&\cos{\theta_{15}\Delta t}& -\sin{\theta_{15}\Delta t}\\
0&0&0&0&\cdots&\sin{\theta_{15}\Delta t}&\cos{\theta_{15}\Delta t}
\end{pmatrix}
\begin{pmatrix}k^{(0)}\\k^{(1)}\\k^{(2)}\\k^{(3)}\\\vdots\\k^{(30)}\\k^{(31)}\end{pmatrix}
\endgroup}_\text{\normalsize modeling temporal dependency with higher frequency} \\
+ \underbrace{\begingroup
\setlength\arraycolsep{1pt}
\begin{pmatrix}q^{(32)}\\q^{(33)}\\q^{(34)}\\q^{(35)}\\\vdots\\q^{(78)}\\q^{(79)}\end{pmatrix}^\top
\begin{pmatrix}
\cos{\theta_{16}\Delta x}& -\sin{\theta_{16}\Delta x}&0&0&\cdots&0&0\\
\sin{\theta_{16}\Delta x}&\cos{\theta_{16}\Delta x}&0&0&\cdots&0&0 \\
0&0&\cos{\theta_{17}\Delta x}& -\sin{\theta_{17}\Delta x}&\cdots&0&0\\
0&0&\sin{\theta_{17}\Delta x}&\cos{\theta_{17}\Delta x}&\cdots&0&0 \\ 
\vdots&\vdots&\vdots&\vdots&\ddots&\vdots&\vdots\\
0&0&0&0&\cdots&\cos{\theta_{39}\Delta x}& -\sin{\theta_{39}\Delta x}\\
0&0&0&0&\cdots&\sin{\theta_{39}\Delta x}&\cos{\theta_{39}\Delta x}
\end{pmatrix}
\begin{pmatrix}k^{(32)}\\k^{(33)}\\k^{(34)}\\k^{(35)}\\\vdots\\k^{(78)}\\k^{(79)}\end{pmatrix}
\endgroup}_\text{\normalsize modeling horizontal dependency with intermediate frequency} \\
+ \underbrace{\begingroup
\setlength\arraycolsep{1pt}
\begin{pmatrix}q^{(80)}\\q^{(81)}\\q^{(82)}\\q^{(83)}\\\vdots\\q^{(126)}\\q^{(127)}\end{pmatrix}^\top
\begin{pmatrix}
\cos{\theta_{40}\Delta y}& -\sin{\theta_{40}\Delta y}&0&0&\cdots&0&0\\
\sin{\theta_{40}\Delta y}&\cos{\theta_{40}\Delta y}&0&0&\cdots&0&0 \\
0&0&\cos{\theta_{41}\Delta y}& -\sin{\theta_{41}\Delta y}&\cdots&0&0\\
0&0&\sin{\theta_{41}\Delta y}&\cos{\theta_{41}\Delta y}&\cdots&0&0 \\ 
\vdots&\vdots&\vdots&\vdots&\ddots&\vdots&\vdots\\
0&0&0&0&\cdots&\cos{\theta_{63}\Delta y}& -\sin{\theta_{63}\Delta y}\\
0&0&0&0&\cdots&\sin{\theta_{63}\Delta y}&\cos{\theta_{63}\Delta y}
\end{pmatrix}
\begin{pmatrix}k^{(80)}\\k^{(81)}\\k^{(82)}\\k^{(83)}\\\vdots\\k^{(126)}\\k^{(127)}\end{pmatrix}
\endgroup}_\text{\normalsize modeling vertical dependency with lower frequency}
\end{gathered}
$}
\label{equ:mrope}
\end{equation}

\noindent \textbf{Frequency Allocation.}
Incorporating 3D structure raises the question of how to allocate the temporal ($t$), horizontal ($x$), and vertical ($y$) components within the $d$ dimensions.
Note that different allocation strategies are not equivalent in the rotation frequency $\theta_{n} = \beta^{-2n/d}$.
As shown in Eq. (\ref{equ:mrope}), M-RoPE assigns higher frequencies (corresponding to lower dimensions) to the temporal dimension ($t$).

To highlight the importance of frequency allocation, we introduce a challenging retrieval task \textbf{V}isual \textbf{N}eedle-\textbf{I}n-\textbf{A}-\textbf{H}astack-\textbf{D}istractor (\textbf{V-NIAH-D}).
V-NIAH-D builds upon V-NIAH \cite{zhang2024longva}, a benchmark designed to evaluate visual long-context understanding.
However, the straightforward retrieval-based task has been shown to provide only a superficial form of long-context understanding~\cite{hsieh2024ruler,yuan2024lv}.
Therefore, We enhance V-NIAH by incorporating semantically similar distractors, obtained using Google Image Search~\cite{googleimagesearch} or Flux ~\cite{flux2023}, to mitigate the possibility of correct answers through random chance.
These distractors are designed to be unambiguous to the question in Fig. \ref{fig:v-ruler}.

\begin{figure}[t]
\centering
\includegraphics[width=.94\linewidth]{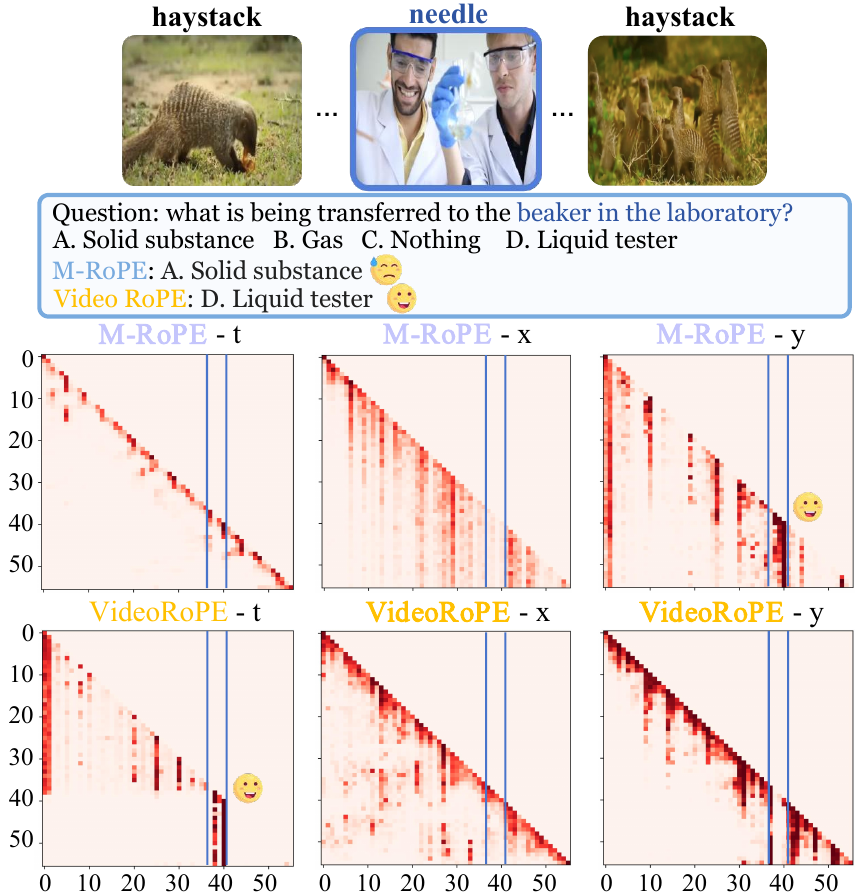}
\vspace{-6pt}
\caption{\footnotesize Attention-based frequential allocation analysis.
\textbf{Middle}: M-RoPE's temporal dimension ($t$) is limited to local information, resulting in a diagonal layout.
\textbf{Bottom}: \methodname effectively retrieves the needle using the temporal dimension.
The x and y coordinates represent the video frame number, e.g., 50 for 50 frames.
For more details see Appendix \ref{app:attention_analysis}.
}
\vspace{-12pt}
\label{fig:attention_analysis}
\end{figure}

As shown in Fig. \ref{fig:v-ruler}, M-RoPE exhibits a clear performance drop from V-NIAH to V-NIAH-D. To investigate this decline, we follow previous works \citep{xiao2023efficient,liu2023scaling,barbero2024round} to visualize the attention scores in Fig. \ref{fig:attention_analysis}. We decompose the attention scores into their corresponding temporal ($t$), horizontal ($x$), and vertical ($y$) components for visualization.

Fig.~\ref{fig:attention_analysis} reveals unusual M-RoPE's attention patterns, despite locating the needle image, it fails to answer the multi-choice question. According to M-RoPE's attention, the needle is located primarily through vertical positional information, rather than temporal features. Thus, the temporal dimension fails to capture long-range semantic dependencies, focusing on local relationships. Conversely, the spatial dimensions capture long-range rather than local semantic information. Lastly, the horizontal and vertical dimensions display distinct characteristics, with the vertical dimension exhibiting phenomena reminiscent of attention sinks \cite{xiao2023efficient}. These suggest the performance decline primarily results from sub-optimal frequency allocation designs of M-RoPE.

\noindent \textbf{Spatial Symmetry.} Given the text tokens $T$ and the visual tokens $T_v$, spatial symmetry \cite{kexuefm10352} claims that the distance between the end of the preceding textual input ($T_{\text{pre}}$) and the beginning of the visual input ($T_v^{\text{start}}$) is equal to the distance between the end of the visual input ($T_v^{\text{end}}$) and the beginning of the subsequent textual input ($T_{\text{sub}}$):
\begin{equation}
    T_{v}^{\text{start}} - T_{\text{pre}} =
    T_{\text{sub}} - T_{v}^{\text{end}}.
\end{equation}
The spatial symmetrical structure can potentially simplify the learning process and reduce bias toward input order.
However, existing 3D RoPE variants such as M-RoPE do not meet the spatial symmetry, we will elaborate related discussion in Fig. \ref{fig:spatial}.

\begin{figure*}[t]
\begin{minipage}{0.98\textwidth}
    \begin{subfigure}[b]{0.49\linewidth}
        \centering
\includegraphics[width=0.95\linewidth]{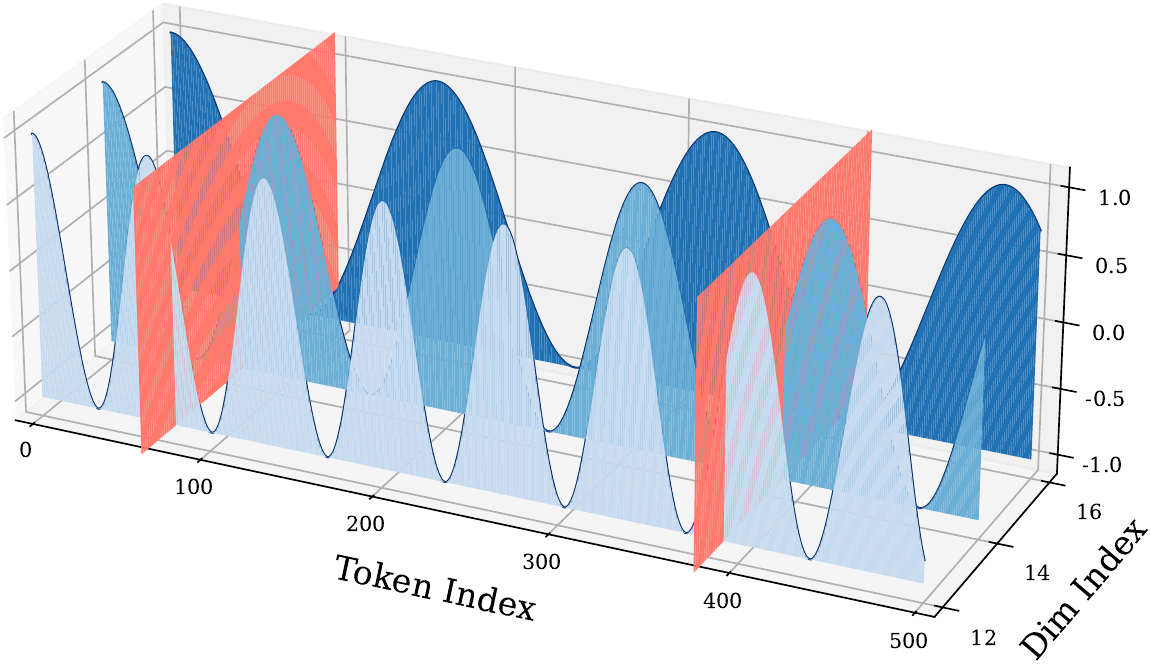}
        \caption{Temporal Frequency Allocation in M-RoPE}
        \label{fig:temporal_mrope}
    \end{subfigure}
    \hfill
    \begin{subfigure}[b]{0.49\linewidth}
        \centering
        \includegraphics[width=0.95\linewidth]{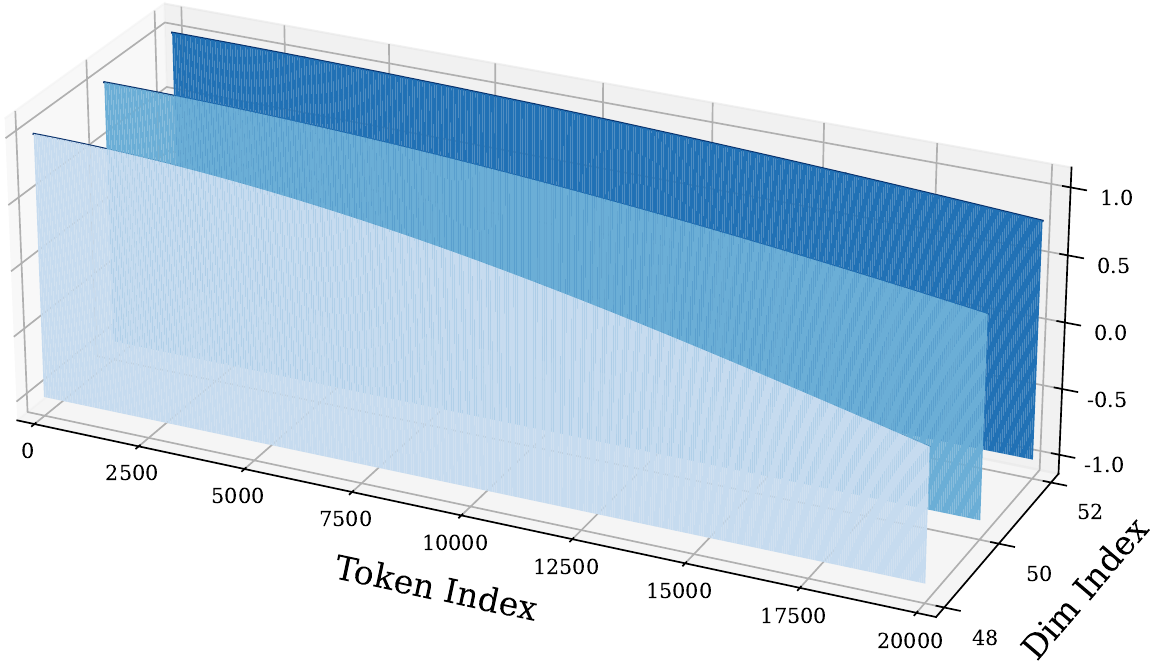}
        \caption{Temporal Frequency Allocation in \methodname (ours)}
        \label{fig:temporal_videorope}
    \end{subfigure}
    \vspace{-6pt}
    \caption{\footnotesize \textbf{(a)} M-RoPE \cite{wang2024qwen2} models temporal dependencies using the \textit{first} 16 rotary angles, which exhibit higher frequencies and more pronounced oscillations. \textbf{(b)} In contrast, \methodname models temporal dependencies using the \textit{last} 16 rotary angles, characterized by significantly wider, monotonic intervals. Our frequency allocation effectively mitigates the misleading influence of distractors in V-NIAH-D. For a more detailed analysis, please refer to Appendix \ref{app:supp_explain_modules}.
    }
    \label{fig:period_mono}
    \vspace{-12pt}
\end{minipage}
\end{figure*}

\noindent \textbf{Temporal Index Scaling.}
The frame index in video and the token index in text are inherently different \cite{kexuefm10352,li2024temporal}.
Recognizing this difference, methods like TAD-RoPE, a 1D RoPE adaptation for Video LLMs, introduce distinct step offsets for image and text token indices: $\gamma$ for image tokens and $\gamma+1$ for text tokens.
Consequently, an ideal RoPE design for video data should permit scaling of the temporal index to meet the inherent difference between the frame index and the text index.

\section{\methodname}\label{subsec:step_size}

Based on some previous research and the above analysis, we claim that a good RoPE design for Video LLMs, especially for long videos, should satisfy four requirements.
The first requirement has been solved by RoPE-Tie~\cite{kexuefm10040} and the subsequent M-RoPE~\cite{wang2024qwen2}.
To solve the last three requirements and mitigate the performance decline observed in V-NIAH-D, we propose our \methodname, comprising the following three key components.

\noindent \textbf{Low-frequency Temporal Allocation (LTA).} 
As shown in Eq. (\ref{equ:rope}), the vanilla RoPE~\cite{su2024roformer} uses all dimensions to model the 1D position information. And as indicated in Eq. (\ref{equ:mrope}), M-RoPE~\cite{wang2024qwen2} uses different dimensions to model temporal, horizontal, and vertical dimensions sequentially.
However, previous frequency allocation strategies are suboptimal because different RoPE dimensions capture dependencies at varying ranges.
As shown in Fig. \ref{fig:attention_analysis}, an interesting observation is that the local attention branch (as reported in \cite{han2024lm}) corresponds to lower dimensions, while the global branch (or attention sink, as in \cite{xiao2023efficient}) corresponds to higher dimensions.
To sum up, lower dimensions (higher frequency, shorter monotonic intervals, larger $\theta_n$) tend to capture relative distances and local semantics \cite{men2024base,barbero2024round}, while higher dimensions (lower frequency, wider monotonic intervals, smaller $\theta_n$) capture longer-range dependencies \cite{barbero2024round}.

Based on our analysis, \methodname uses higher dimensions for temporal features in longer contexts and lower dimensions for spatial features, which are limited by resolution and have a fixed range.
To avoid the gap between horizontal and vertical positions, we interleave the dimensions responsible for these spatial features.
The dimension distribution for \methodname is shown in Eq. (\ref{equ:videorope}):


\begin{equation}
\resizebox{0.5\textwidth}{!}{$
\scriptsize
\begin{gathered}
\underbrace{\begingroup
\setlength\arraycolsep{1pt}
\begin{pmatrix}q^{(96)}\\q^{(97)}\\q^{(98)}\\q^{(99)}\\\vdots\\q^{(126)}\\q^{(127)}\end{pmatrix}^\top
\begin{pmatrix}
\cos{\theta_{48}\Delta t}& -\sin{\theta_{48}\Delta t}&0&0&\cdots&0&0\\
\sin{\theta_{48}\Delta t}&\cos{\theta_{48}\Delta t}&0&0&\cdots&0&0 \\
0&0&\cos{\theta_{49}\Delta t}& -\sin{\theta_{49}\Delta t}&\cdots&0&0\\
0&0&\sin{\theta_{49}\Delta t}&\cos{\theta_{49}\Delta t}&\cdots&0&0 \\ 
\vdots&\vdots&\vdots&\vdots&\ddots&\vdots&\vdots\\
0&0&0&0&\cdots&\cos{\theta_{63}\Delta t}& -\sin{\theta_{63}\Delta t}\\
0&0&0&0&\cdots&\sin{\theta_{63}\Delta t}&\cos{\theta_{63}\Delta t}
\end{pmatrix}
\begin{pmatrix}k^{(96)}\\k^{(97)}\\k^{(98)}\\k^{(99)}\\\vdots\\k^{(126)}\\k^{(127)}\end{pmatrix}
\endgroup}_\text{\normalsize modeling temporal dependency with lower frequency} \\
+ \underbrace{\begingroup
\setlength\arraycolsep{1pt}
\begin{pmatrix}q^{(0)}\\q^{(1)}\\q^{(4)}\\q^{(5)}\\\vdots\\q^{(92)}\\q^{(93)}\end{pmatrix}^\top
\begin{pmatrix}
\cos{\theta_{0}\Delta x}& -\sin{\theta_{0}\Delta x}&0&0&\cdots&0&0\\
\sin{\theta_{0}\Delta x}&\cos{\theta_{0}\Delta x}&0&0&\cdots&0&0 \\
0&0&\cos{\theta_{2}\Delta x}& -\sin{\theta_{2}\Delta x}&\cdots&0&0\\
0&0&\sin{\theta_{2}\Delta x}&\cos{\theta_{2}\Delta x}&\cdots&0&0 \\ 
\vdots&\vdots&\vdots&\vdots&\ddots&\vdots&\vdots\\
0&0&0&0&\cdots&\cos{\theta_{46}\Delta x}& -\sin{\theta_{46}\Delta x}\\
0&0&0&0&\cdots&\sin{\theta_{46}\Delta x}&\cos{\theta_{46}\Delta x}
\end{pmatrix}
\begin{pmatrix}k^{(0)}\\k^{(1)}\\k^{(4)}\\k^{(5)}\\\vdots\\k^{(92)}\\k^{(93)}\end{pmatrix}
\endgroup}_\text{\normalsize modeling horizontal dependency with interleaved high frequency} \\
+ \underbrace{\begingroup
\setlength\arraycolsep{1pt}
\begin{pmatrix}q^{(2)}\\q^{(3)}\\q^{(6)}\\q^{(7)}\\\vdots\\q^{(94)}\\q^{(95)}\end{pmatrix}^\top
\begin{pmatrix}
\cos{\theta_{1}\Delta y}& -\sin{\theta_{1}\Delta y}&0&0&\cdots&0&0\\
\sin{\theta_{1}\Delta y}&\cos{\theta_{1}\Delta y}&0&0&\cdots&0&0 \\
0&0&\cos{\theta_{3}\Delta y}& -\sin{\theta_{3}\Delta y}&\cdots&0&0\\
0&0&\sin{\theta_{3}\Delta y}&\cos{\theta_{3}\Delta y}&\cdots&0&0 \\ 
\vdots&\vdots&\vdots&\vdots&\ddots&\vdots&\vdots\\
0&0&0&0&\cdots&\cos{\theta_{47}\Delta y}& -\sin{\theta_{47}\Delta y}\\
0&0&0&0&\cdots&\sin{\theta_{47}\Delta y}&\cos{\theta_{47}\Delta y}
\end{pmatrix}
\begin{pmatrix}k^{(2)}\\k^{(3)}\\k^{(6)}\\k^{(7)}\\\vdots\\k^{(94)}\\k^{(95)}\end{pmatrix}
\endgroup}_\text{\normalsize modeling vertical dependency with interleaved high frequency} \\
\end{gathered}
$
}
\label{equ:videorope}
\end{equation}
The horizontal position $x$ and vertical position $y$ are interleaved to occupy the lower dimensions, followed by temporal $t$, which occupies the higher dimensions. We keep the same allocation number for $x$, $y$, and $t$ as M-RoPE for a fair comparison, with values of 48, 48, and 32, respectively.
The advantages of this distribution are evident in Fig.  \ref{fig:period_mono}. 
For a RoPE-based LLM with a 128-dimensional head (64 rotary angles $\theta_n$), we visualize the function of $\cos{\theta_n t}$ for 3 dimensions using parallel blue planes.

As shown in Fig. \ref{fig:period_mono} (\textbf{a}), M-RoPE's temporal position embeddings are significantly distorted by periodic oscillations \cite{men2024base}, leading to identical embeddings for distant positions.
For instance, considering the last three rotary angles, the temporal embeddings are severely affected by these oscillations due to their short monotonic intervals (and even shorter intervals in lower dimensions).
This periodicity creates ``hash collisions'' (red planes), where distant positions share near-identical embeddings, making the model susceptible to distractor influence.
Fortunately, our \methodname (Fig. \ref{fig:period_mono} (\textbf{b})) is free from oscillation and Hash collision in temporal modeling.
The relationship between periodicity, monotonicity, and temporal modeling is visualized in Fig \ref{fig:period_mono}.

\begin{figure}[t]
\centering
\includegraphics[width=0.98\linewidth]{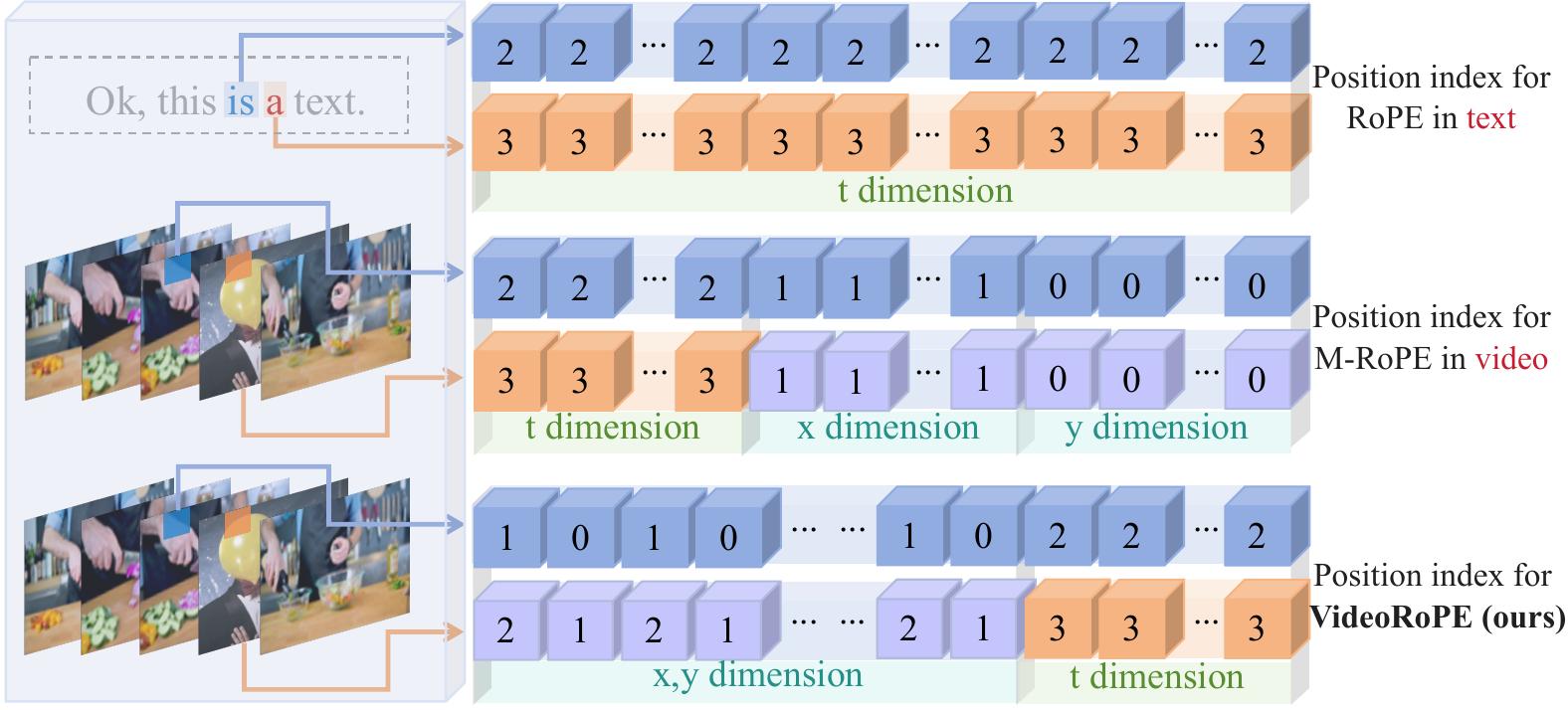}
\vspace{-6pt}
\caption{\footnotesize The position embeddings of adjacent text tokens for Vanilla RoPE (\textbf{top} row), the corresponding visual tokens in adjacent frames for M-RoPE (\textbf{middle} row) and our \methodname (\textbf{bottom} row) with interleaved spatial and temporal last design.}
\vspace{-12pt}
\label{fig:spatail_index}
\end{figure}

\begin{figure*}[t]
\begin{minipage}{0.98\textwidth}
    \begin{subfigure}[b]{0.3\linewidth}
        \centering
        \includegraphics[width=0.95\linewidth]{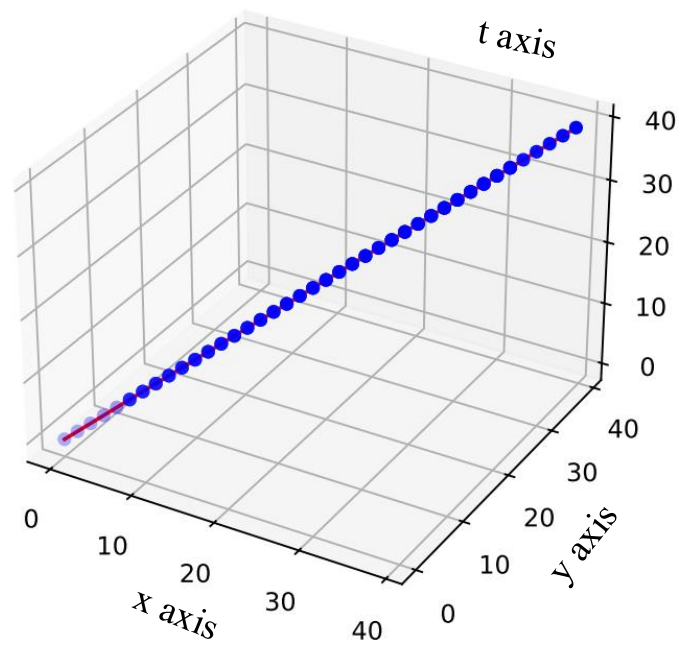}
        \caption{3D visualization for Vanilla RoPE.}
        \label{fig:vanilla_rope}
    \end{subfigure}
    \hfill
    \begin{subfigure}[b]{0.3\linewidth}
        \centering
        \includegraphics[width=0.95\linewidth]{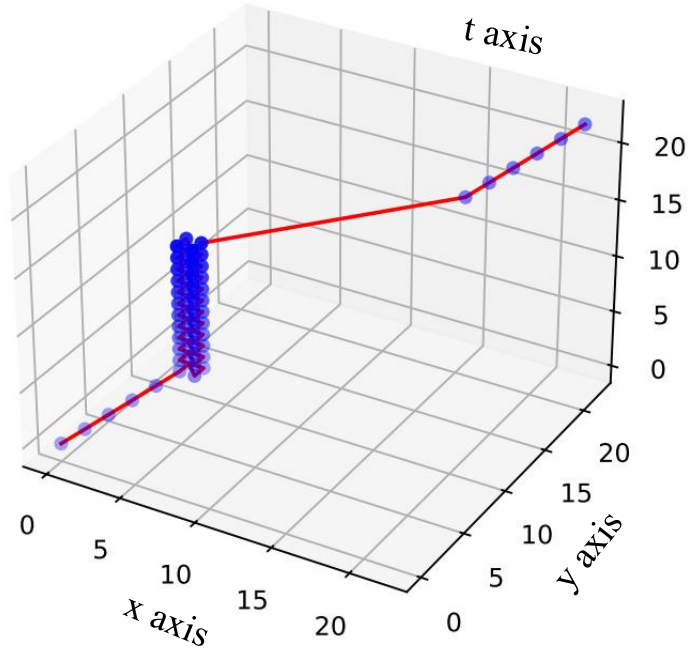}
        \caption{3D visualization for M-RoPE.}
        \label{fig:m_rope}
    \end{subfigure}
    \hfill
    \begin{subfigure}[b]{0.3\linewidth}
        \centering
        \includegraphics[width=0.95\linewidth]{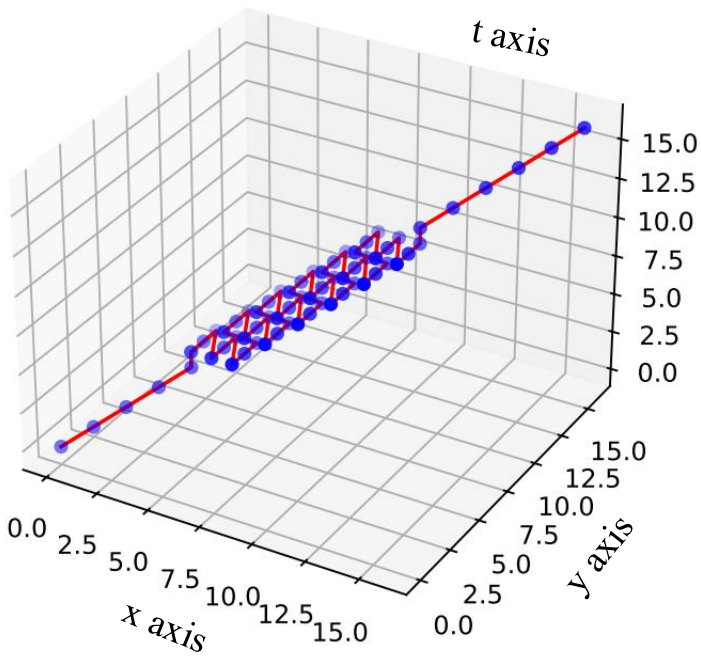}
        \caption{3D visualization for \methodname.}
        \label{fig:video_rope}
    \end{subfigure}
    \hfill
    \vspace{-6pt}
    \caption{\footnotesize The 3D visualization for different position embedding. \textbf{(a)} The vanilla 1D RoPE~\cite{su2024roformer} does not incorporate spatial modeling.
    \textbf{(b)} M-RoPE~\cite{wang2024qwen2}, while have the 3D structure, introduces a discrepancy in index growth for visual tokens across frames, with some indices remaining constant.
    \textbf{(c)} In contrast, our \methodname achieves the desired balance, maintaining the consistent index growth pattern of vanilla RoPE while simultaneously incorporating spatial modeling. 
    }
    \vspace{-12pt}
    \label{fig:spatial}
\end{minipage}
\end{figure*}

\noindent \textbf{Diagonal Layout.}
Fig. \ref{fig:spatial} provides a visual comparison of spatial symmetry in positional encodings.
For vanilla RoPE (Fig. ~\ref{fig:vanilla_rope}), no spatial relation is considered and the index for every dimension increases directly.
While M-RoPE (Fig. \ref{fig:m_rope}), incorporates spatial information within each frame, it introduces two significant discontinuities between textual and visual tokens.
This arises from M-RoPE's placement strategy, if the first visual token is at $(0, 0)$, the last token in each frame will always be placed at $(W-1, H-1)$, creating a stack in the bottom-left corner.
Furthermore, like vanilla RoPE, M-RoPE's indices increase with input length across all dimensions.

To address these limitations, \methodname arranges the entire input along the diagonal, see Fig. \ref{fig:video_rope}.
The central patch's 3D position for each video frame is $(t,t,t)$, with other patches offset in all directions.
Our \textbf{Diagonal Layout} has two advantages: (1) our design preserves the relative positions of visual tokens and ensures approximate equidistance from the image corners to the center, preventing text tokens from being overly close to any corner. (2) It maintains the indexing pattern of vanilla RoPE (Fig.  \ref{fig:spatail_index}), as the position index increment between corresponding spatial locations in adjacent frames mirrors that of adjacent textual tokens.

\noindent \textbf{Adjustable Temporal Spacing.}
To scale the temporal index, we introduce a scaling factor $\delta$ to better align temporal information between visual and textual tokens.

Suppose the symbol $\tau$ denotes the token index, for the starting text ($0 \leq \tau < T_s$), the temporal, horizontal, and vertical indices are simply set to the raw token index $\tau$.
For the video input ($T_s \leq \tau < T_s + T_v$), The difference $\tau - T_s$ represents the index of the current frame relative to the start of the video, which is then scaled by $\delta$ to control the space in the temporal dimension.
For the ending text ($T_s + T_v \leq \tau < T_s + T_v + T_e$), the temporal, horizontal, and vertical index are the same, creating a linear progression.

According to our adjustable temporal spacing design, for a multi-modal input that consists of a text with $T_s$ tokens, a following video with $T_v$ frame with $W\times H$ patches in each frame, and an ending text with $T_e$ tokens, the position indices $(t, x, y)$ of \methodname for $\tau$-th textual token or $(\tau, w, h)$-th visual token are defined as Eq. (\ref{equ:index}):
\begin{equation}
\vspace{-3pt}
\resizebox{0.5\textwidth}{!}{$
    \footnotesize
    (t,x,y) =
    \begin{cases}
        (\tau, \tau, \tau) & \text{if } 0 \leq \tau < T_s \\[3ex]
        \left( 
        \begin{array}{l}
            T_s + \delta (\tau - T_s), \\
            T_s + \delta (\tau - T_s) + w - \frac{W}{2}, \\
            T_s + \delta (\tau - T_s) + h - \frac{H}{2}
        \end{array}
        \right) & \text{if } T_s \leq \tau < T_s + T_v \\[6ex]
        \left( 
        \begin{array}{l}
            \tau + (\delta-1) T_v, \\
            \tau + (\delta-1) T_v, \\
            \tau + (\delta-1) T_v
        \end{array}
        \right) & \text{if } T_s + T_v \leq \tau < T_s + T_v + T_e
    \end{cases}
$}
\raisebox{-9.5ex}{,}
\label{equ:index}
\end{equation}
where $w$ and $h$ represent the horizontal and vertical indices of the visual patch within the frame, respectively.  

In summary, the parameter $\delta$ in our adjustable temporal spacing allows for a flexible and consistent way to encode the relative positions of text and video tokens.
\section{Experiment}

\renewcommand{\arraystretch}{1.1}
\begin{table*}[!ht]
\setlength\tabcolsep{5pt} 
\centering
\caption{\textbf{Comparison of different RoPE methods on LongVidionBench, MLVU, and Video-MME}. The benchmarks evaluate performance across three context lengths: 8k, 16k, 32k, and 64k, where \textbf{8k} represents context within the training range, and others represent context outside the training range. Our \methodname outperforms other RoPE variants across all three benchmarks. The best results are marked in \textbf{bold}, and the second-best results are \underline{underlined}. For more information on the evaluation, see Appendix \ref{appendix:benchmarks}.}
\label{tab:lvlm_all}
\vspace{2mm}
\footnotesize
\begin{tabular}{cllllllllllll}
\toprule
\multirow{2}{*}{\textbf{Method}} & \multicolumn{4}{c}{\textbf{LongVideoBench}} & \multicolumn{4}{c}{\textbf{MLVU}} & \multicolumn{4}{c}{\textbf{Video-MME}} \\ 
\cmidrule(lr){2-5} 
\cmidrule(lr){6-9} 
\cmidrule(lr){10-13} 
 & 8k & 16k & 32k & 64k & 8k & 16k & 32k & 64k & 8k & 16k & 32k & 64k \\ 
\hline
Vanilla RoPE \cite{su2024roformer} & \textbf{54.97} & 54.87 & \underline{54.56} & 54.04 & 63.31 & \underline{65.79} &\underline{65.93} & \underline{62.02} & \underline{60.67} & 60.00 & 61.33 & 58.33 \\
TAD-RoPE \cite{gao2024tc} & 54.14 & \underline{55.08} & 53.94 & 53.42 & \underline{63.67} & 65.28 & 65.28 & 60.73 & 60.33 & \textbf{61.33} & \textbf{62.00} & 58.67 \\
M-RoPE \cite{wang2024qwen2} & 53.42 & 52.80 & 53.11 & \underline{54.35} & 60.41 & 60.68 & 61.56 & 61.10 & \underline{60.67} & 59.67 & 61.00 & \underline{59.67} \\
\hline
\rowcolor[HTML]{F2F3F5}
\methodname (Ours) & \underline{54.46} & \textbf{55.29} & \textbf{57.15} & \textbf{57.26} & \textbf{65.19} & \textbf{66.29} & \textbf{66.02} & \textbf{65.56} & \textbf{61.33} & \underline{61.00} &\underline{61.67} & \textbf{61.33} \\
\bottomrule
\end{tabular}
\end{table*}

\begin{figure*}[ht]
\centering
\includegraphics[width=0.99\linewidth]{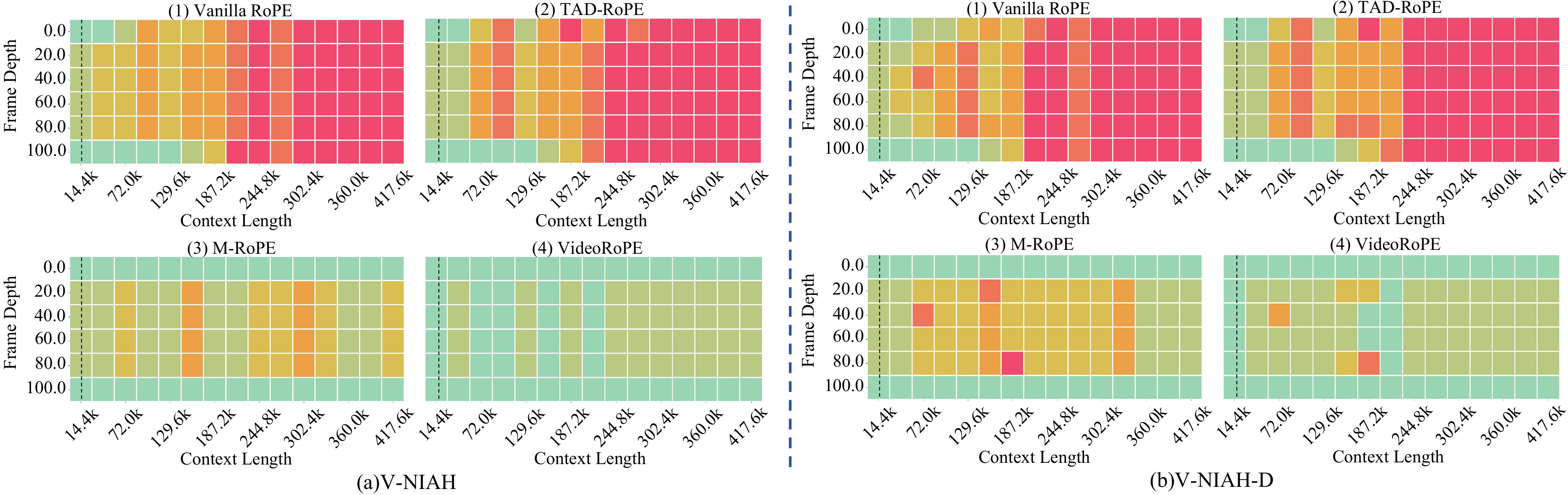}
\vspace{-6pt}
\caption{Visualization of the retrieval results for V-NIAH and V-NIAH-D. The color gradient from green to red represents the progression of needle retrieval performance, from perfect to zero.}
\vspace{-12pt}
\label{fig:v-niah-and-d}
\end{figure*}

\subsection{Experimental Setup}


\noindent \textbf{Training Data.} We use a subset of LLaVA-Video-178k dataset \cite{zhang2024video} to train \methodname.
The LLaVA-Video-178k dataset covers 178k videos and around 5 million question-answers (QA) pairs from diverse sources such as HD-VILA \cite{xue2022hdvila}, Kinetics \cite{kay2017kinetics}, and ActivityNet \cite{caba2015activitynet}.
To balance training efficiency and long-video comprehension, we randomly select 136k videos with durations under 2 minutes and 18k videos with durations between 2 and 3 minutes.
This process yielded our training set of approximately 1.3 million pairs.

\noindent \textbf{Implementation Details.}
Using the aforementioned video training data, we fine-tune different models that use different positional encoding strategies, such as the Vanilla RoPE \cite{su2024roformer}, Time-Aware Dual RoPE (TAD-RoPE) \cite{gao2024tc}, M-RoPE \cite{wang2024qwen2}, and our \methodname.
All models are initialized with the Vision Transformer from Qwen2-VL-7B and LLM (Vanilla RoPE) from Qwen2-7B \cite{yang2024qwen2}.
Our fine-tuning incorporates our \methodname to process the spatiotemporal nature of the video data effectively.
We adopt Qwen2-VL's fine-tuning settings, processing each video at 2 fps with a maximum of 128 frames and dynamically adjusting the image resolution to maintain a consistent token count. However, to prevent memory overflow, we use a context window of 8192 tokens.

Our fine-tuning process employs a batch size of 128, a cosine scheduler with a learning rate of 1e-5, a warm-up ratio of 1e-2, and 704 Nvidia-A100 GPU hours in total.
The evaluation involves sampling videos at 2 fps with a minimum of 144 image tokens per frame.
We use the vLLM framework \cite{kwon2023efficient} to support inference on sequences longer than 32k tokens.

\noindent \textbf{Evaluation Benchmarks.} We evaluate our approach using six video benchmarks, including tasks related to \textit{long video understanding}, \textit{long video retrieval}, and \textit{video hallucination}. For \textit{long video understanding}, we use \textbf{LongVideoBench} \cite{wu2024longvideobench} (8 seconds to 1 hour), \textbf{MLVU} \cite{zhou2024mlvu} (3 minutes to 2 hours), and \textbf{Video-MME} \cite{fu2024video} (11 seconds to 60 minutes). For \textit{long video retrieval}, we use \textbf{Vision Needle-in-a-Haystack (V-NIAH)} \cite{zhang2024longva} and our proposed extension, \textbf{Vision Needle-in-a-Haystack with Distractors (V-NIAH-D)}, which introduces distractor frames to increase the task difficulty. For \textit{video hallucination}, we use \textbf{VideoHallucer} \cite{videohallucer}, which evaluates the model's ability to correctly answer both basic and hallucinated questions about video content. Details of these benchmarks can be found in  Appendix~\ref{appendix:benchmarks}.

\subsection{Results on Long Video Understanding}
As shown in Tab. \ref{tab:lvlm_all}, we compare our \methodname with existing RoPE variants (vanilla RoPE \cite{su2024roformer}, TAD-RoPE \cite{gao2024tc}, and M-RoPE \cite{wang2024qwen2}) across three prominent video understanding benchmarks. Our \methodname consistently outperforms all baseline methods across these benchmarks, demonstrating its robustness and adaptability. Specifically, \methodname achieves improvements of up to 2.91, 4.46, and 1.66 points (64k context length) over the M-RoPE baseline on LongVideoBench, MLVU, and Video-MME, respectively. These results emphasize the superior ability of \methodname to effectively capture long-range dependencies and maintain performance across various challenging video data tasks.

\renewcommand{\arraystretch}{1.}
\setlength\tabcolsep{2pt}
\setlength{\textfloatsep}{5pt}
\begin{table}[t]
\centering
\footnotesize
\caption{
\textbf{Performance comparison of different RoPEs on V-NIAH and V-NIAH-D}. ``Acc.'' refers to the average accuracy across haystack length and frame depth.
}
\label{tab:v-niah-and-d}
\vspace{2mm}
\begin{tabular}{ccc}
\toprule
\textbf{Method} & \textbf{V-NIAH Acc.} & \textbf{V-NIAH-D Acc.} \\ \hline
Vanilla RoPE~\cite{su2024roformer} & 31.78 & 30.22 \\
TAD-RoPE~\cite{gao2024tc} & 29.33 & 29.56 \\
M-RoPE~\cite{wang2024qwen2} & \underline{78.67} & \underline{74.67} \\
\hline
\rowcolor[HTML]{F2F3F5}
\methodname & \textbf{91.11}  &\textbf{87.11} \\
\bottomrule
\end{tabular}
\end{table}
\setlength{\textfloatsep}{1pt}  
\setlength{\intextsep}{1pt}     
\setlength{\floatsep}{1pt} 
\renewcommand{\arraystretch}{1.0}
\setlength{\tabcolsep}{2pt}
\setlength{\textfloatsep}{5pt}
\begin{table}[t]
\centering
\footnotesize
\caption {\textbf{Performance comparison of different RoPEs on VideoHallucer}, evaluated at context lengths of 8k, 16k, 32k, and 64k. The maximum result for each RoPE variant across these context lengths is displayed, with bold for the top result and underlined for the second-highest. `OR' = Object-Relation, `T' = Temporal, `SD' = Semantic Detail, `F' = Factual, `NF' = Non-factual.
}
\label{tab:video_hallucer}
\vspace{2mm}
\begin{tabular}{ccccccc}
\toprule
\textbf{Method} & \textbf{OR} & \textbf{T} & \textbf{SD} & \textbf{F} & \textbf{NF} & \textbf{Avg.} \\ \hline
Vanilla RoPE~\cite{su2024roformer} & \underline{51.5} & 30.0 & \underline{48.0} & 8.0 & 43.0 & 36.1 \\
TAD-RoPE~\cite{gao2024tc} & 51.0 & \underline{37.0} & \underline{48.0} & 11.5 & \underline{47.5} & 39.0 \\
M-RoPE~\cite{wang2024qwen2} & 39.0 & 29.0 & 43.5 & \underline{12.5} & \underline{47.5} & 34.3 \\
\hline
\rowcolor[HTML]{F2F3F5}
\methodname & \textbf{57.0} & \textbf{58.5} & \textbf{50.5} & \textbf{15.0} & \textbf{50.0} & \textbf{46.2} \\
\bottomrule
\end{tabular}
\end{table}
\subsection{Results on Long Video Retrieval}
Fig. \ref{fig:v-niah-and-d} illustrates the performance of V-NIAH and V-NIAH-D with \methodname and other RoPE variants. Specifically, Fig. \ref{fig:v-niah-and-d} (a) and (b) demonstrate that the proposed V-NIAH-D is more challenging than V-NIAH.
Fig. \ref{fig:v-niah-and-d} (1) and (2) show that both Vanilla RoPE and TAD-RoPE exhibit some extrapolation ability beyond the visual training context. However, both methods fail once they exceed a certain extrapolation limit.
In contrast, Fig. \ref{fig:v-niah-and-d} (3) and (4) highlight the superior performance of \methodname and M-RoPE in extrapolating within the test context range. While both \methodname and M-RoPE successfully handle extrapolation, \methodname consistently outperforms M-RoPE, showcasing the robustness of the task.
Tab. \ref{tab:v-niah-and-d} provides a quantitative analysis of the retrieval results, demonstrating a 12.44
\% performance improvement of our method over M-RoPE on the Video Retrieval task in both settings, confirming the advantages of our proposed method in video retrieval scenarios.

\subsection{Results on Video Hallucination}

As highlighted in Tab. \ref{tab:video_hallucer}, \methodname significantly surpasses current RoPE methods on the VideoHallucer benchmark. In particular, for the Temporal Hallucination task, \methodname demonstrates a substantial performance improvement of 29.5\%, indicating its enhanced capability to accurately capture and process temporal dependencies. This improvement suggests that \methodname is better equipped to handle dynamic video sequences, where the understanding of time-based relationships is critical. Similarly, for the Spatial Hallucination task, specifically the Object-Relation Hallucination subtask, \methodname achieves an impressive 18.0\% improvement over existing methods, highlighting its ability to better discern complex spatial interactions. These results underscore \methodname's robustness in solving video hallucination and potential for real-world video analysis.

\subsection{Ablation Studies}
\noindent \textbf{Ablation Studies on Module Design.} 

We conduct ablation experiments on the modules introduced in Section \ref{subsec:step_size}, quantitatively evaluating their impact on LongVideoBench and MLVU benchmarks. The experimental results are presented in Tab. \ref{tab:ablation_modules}. The baseline setting, M-RoPE~\cite{wang2024qwen2}, achieves scores of 54.35 on LongVideoBench and 61.10 on MLVU (both using a 64k context length). By progressively integrating the DL (Diagonal Layout), LTA (Low-frequency Temporal Allocation), and ATS (Adjustable Temporal Spacing) modules, our method shows a continuous improvement in performance, achieving enhanced scores of 57.26 on LongVideoBench and 65.56 on MLVU (both using a 64k context length). These results demonstrate the effectiveness of our approach in leveraging spatial-temporal positional information. 
To refine the analysis of \( x \) and \( y \) allocation in LTA, we quantitatively evaluate interleaved vs.\ sequential layouts. We also compare strategies for allocating \( t \), \( x \), and \( y \), including M-RoPE, a uniform interleaved layout, and our VideoRoPE design. Additionally, we explore the optimal ATS scaling factor by varying its value, and further ablate the diagonal layout module to validate the symmetry-based design. See Appendix~\ref{app:ablation_study} for details.

\renewcommand{\arraystretch}{1.0} 
\begin{table}[t]
\setlength\tabcolsep{3pt} 
\centering
\caption {\textbf{Ablation study about different modules of \methodname.}
}\label{tab:ablation_modules}
\vspace{1mm} 
\scriptsize 
\begin{tabular}{@{}clllllllll@{}}
\hline
\multirow{2}{*}{\textbf{Method}}  & \multicolumn{4}{c}{\textbf{LongVideoBench}} & \multicolumn{4}{c}{\textbf{MLVU}} \\ 
\cmidrule(lr){2-5}
\cmidrule(lr){6-9}
 & 8k & 16k & 32k & 64k & 8k & 16k & 32k & 64k \\ \hline
Baseline & 53.42 & 52.80 & 53.11 & 54.35 & 60.41 & 60.68 & 61.56 & 61.10 \\
+ DL & 52.17 & 52.07 & 53.31 & 53.63 & 62.06 & 63.03 & 62.52 & 62.75 \\
+ DL \& LTA & \textbf{54.46} & \textbf{55.49} & 54.66 & 55.60 & 63.35 & 64.09 & 64.00 & 63.26  \\
+ DL \& LTA \& ATS & {\textbf{54.46}} & 55.29 & {\textbf{57.15}} & {\textbf{57.26}} & {\textbf{65.19}} & \textbf{66.29} & \textbf{66.02} & \textbf{65.56} \\ \hline
\end{tabular}
\end{table}

\vspace{-3mm}
\section{Conclusion}
This paper identifies four key criteria for effective positional encoding: 2D/3D structure, frequency allocation, spatial symmetry, and temporal index scaling. As part of our analysis, through the V-NIAH-D task, we demonstrate that previous RoPE variants are vulnerable to distractors because of lacking proper temporal allocation. As a result, We propose \methodname that uses a 3D structure for spatiotemporal coherence, low-frequency temporal allocation to reduce oscillations, a diagonal layout for spatial symmetry, and adjustable temporal spacing. \methodname outperforms previous RoPE variants in tasks like long video retrieval, video understanding, and video hallucination.

\section*{Acknowledgments}
This work was supported by National Key R\&D Program of China 2022ZD0161600, Shanghai Artificial lntelligence Laboratory, the Centre for Perceptual and Interactive Intelligence (CPII) Ltd under the Innovation and Technology Commission (ITC)’s InnoHK. Dahua Lin is a PI of CPII under the InnoHK.

\section*{Impact Statement}
This paper presents work whose goal is to advance the field
of Machine Learning. There are many potential societal
consequences of our work, and none of which we feel must
be specifically highlighted here.

\bibliography{main}

\begin{thebibliography}{88}
\providecommand{\natexlab}[1]{#1}
\providecommand{\url}[1]{\texttt{#1}}
\expandafter\ifx\csname urlstyle\endcsname\relax
  \providecommand{\doi}[1]{doi: #1}\else
  \providecommand{\doi}{doi: \begingroup \urlstyle{rm}\Url}\fi

\bibitem[Agrawal et~al.(2024)Agrawal, Antoniak, Hanna, Bout, Chaplot,
  Chudnovsky, Costa, Monicault, Garg, Gervet, Ghosh, Héliou, Jacob, Jiang,
  Khandelwal, Lacroix, Lample, Casas, Lavril, Scao, Lo, Marshall, Martin,
  Mensch, Muddireddy, Nemychnikova, Pellat, Platen, Raghuraman, Rozière,
  Sablayrolles, Saulnier, Sauvestre, Shang, Soletskyi, Stewart, Stock, Studnia,
  Subramanian, Vaze, Wang, and Yang]{agrawal2024pixtral12b}
Agrawal, P., Antoniak, S., Hanna, E.~B., Bout, B., Chaplot, D., Chudnovsky, J.,
  Costa, D., Monicault, B.~D., Garg, S., Gervet, T., Ghosh, S., Héliou, A.,
  Jacob, P., Jiang, A.~Q., Khandelwal, K., Lacroix, T., Lample, G., Casas,
  D.~L., Lavril, T., Scao, T.~L., Lo, A., Marshall, W., Martin, L., Mensch, A.,
  Muddireddy, P., Nemychnikova, V., Pellat, M., Platen, P.~V., Raghuraman, N.,
  Rozière, B., Sablayrolles, A., Saulnier, L., Sauvestre, R., Shang, W.,
  Soletskyi, R., Stewart, L., Stock, P., Studnia, J., Subramanian, S., Vaze,
  S., Wang, T., and Yang, S.
\newblock Pixtral 12b, 2024.
\newblock URL \url{https://arxiv.org/abs/2410.07073}.

\bibitem[Barbero et~al.(2024)Barbero, Vitvitskyi, Perivolaropoulos, Pascanu,
  and Veli{\v{c}}kovi{\'c}]{barbero2024round}
Barbero, F., Vitvitskyi, A., Perivolaropoulos, C., Pascanu, R., and
  Veli{\v{c}}kovi{\'c}, P.
\newblock Round and round we go! what makes rotary positional encodings useful?
\newblock \emph{arXiv preprint arXiv:2410.06205}, 2024.

\bibitem[Bertasius et~al.(2021)Bertasius, Wang, and
  Torresani]{bertasius2021spacetimeattentionneedvideo}
Bertasius, G., Wang, H., and Torresani, L.
\newblock Is space-time attention all you need for video understanding?, 2021.
\newblock URL \url{https://arxiv.org/abs/2102.05095}.

\bibitem[Cai et~al.(2024)Cai, Cao, Chen, Chen, Chen, Chen, Chen, Chen, Chen,
  Chu, Dong, Duan, Fan, Fei, Gao, Ge, Gu, Gu, Gui, Guo, Guo, He, Hu, Huang,
  Jiang, Jiao, Jin, Lei, Li, Li, Li, Li, Li, Li, Liu, Liu, Hong, Liu, Liu, Liu,
  Lv, Lv, Lv, Ma, Ma, Ma, Ning, Ouyang, Qiu, Qu, Shang, Shao, Song, Song, Sui,
  Sun, Sun, Tang, Wang, Wang, Wang, Wang, Wang, Wang, Wang, Wei, Weng, Wu,
  Xiong, Xu, Xu, Yan, Yan, Yang, Ye, Ying, Yu, Yu, Zang, Zhang, Zhang, Zhang,
  Zhang, Zhang, Zhang, Zhang, Zhang, Zhang, Zhang, Zhang, Zhao, Zhao, Zhao,
  Zhou, Zhou, Zhuo, Zou, Qiu, Qiao, and Lin]{cai2024internlm2}
Cai, Z., Cao, M., Chen, H., Chen, K., Chen, K., Chen, X., Chen, X., Chen, Z.,
  Chen, Z., Chu, P., Dong, X., Duan, H., Fan, Q., Fei, Z., Gao, Y., Ge, J., Gu,
  C., Gu, Y., Gui, T., Guo, A., Guo, Q., He, C., Hu, Y., Huang, T., Jiang, T.,
  Jiao, P., Jin, Z., Lei, Z., Li, J., Li, J., Li, L., Li, S., Li, W., Li, Y.,
  Liu, H., Liu, J., Hong, J., Liu, K., Liu, K., Liu, X., Lv, C., Lv, H., Lv,
  K., Ma, L., Ma, R., Ma, Z., Ning, W., Ouyang, L., Qiu, J., Qu, Y., Shang, F.,
  Shao, Y., Song, D., Song, Z., Sui, Z., Sun, P., Sun, Y., Tang, H., Wang, B.,
  Wang, G., Wang, J., Wang, J., Wang, R., Wang, Y., Wang, Z., Wei, X., Weng,
  Q., Wu, F., Xiong, Y., Xu, C., Xu, R., Yan, H., Yan, Y., Yang, X., Ye, H.,
  Ying, H., Yu, J., Yu, J., Zang, Y., Zhang, C., Zhang, L., Zhang, P., Zhang,
  P., Zhang, R., Zhang, S., Zhang, S., Zhang, W., Zhang, W., Zhang, X., Zhang,
  X., Zhao, H., Zhao, Q., Zhao, X., Zhou, F., Zhou, Z., Zhuo, J., Zou, Y., Qiu,
  X., Qiao, Y., and Lin, D.
\newblock Internlm2 technical report, 2024.

\bibitem[Chai et~al.(2024)Chai, Song, Du, Meng, Madhavan, Bar-Tal, Hwang, Xie,
  and Manning]{chai2024auroracapefficientperformantvideo}
Chai, W., Song, E., Du, Y., Meng, C., Madhavan, V., Bar-Tal, O., Hwang, J.-N.,
  Xie, S., and Manning, C.~D.
\newblock Auroracap: Efficient, performant video detailed captioning and a new
  benchmark, 2024.
\newblock URL \url{https://arxiv.org/abs/2410.03051}.

\bibitem[Chen \& Xing(2024)Chen and Xing]{chen2024open}
Chen, L. and Xing, L.
\newblock Open-llava-next: An open-source implementation of llava-next series
  for facilitating the large multi-modal model community.
\newblock \url{https://github.com/xiaoachen98/Open-LLaVA-NeXT}, 2024.

\bibitem[Chen et~al.(2023)Chen, Li, Dong, Zhang, He, Wang, Zhao, and
  Lin]{chen2023sharegpt4v}
Chen, L., Li, J., Dong, X., Zhang, P., He, C., Wang, J., Zhao, F., and Lin, D.
\newblock Sharegpt4v: Improving large multi-modal models with better captions.
\newblock \emph{arXiv preprint arXiv:2311.12793}, 2023.

\bibitem[Chen et~al.(2024{\natexlab{a}})Chen, Wei, Li, Dong, Zhang, Zang, Chen,
  Duan, Lin, Tang, Yuan, Qiao, Lin, Zhao, and Wang]{chen2024sharegpt4video}
Chen, L., Wei, X., Li, J., Dong, X., Zhang, P., Zang, Y., Chen, Z., Duan, H.,
  Lin, B., Tang, Z., Yuan, L., Qiao, Y., Lin, D., Zhao, F., and Wang, J.
\newblock Sharegpt4video: Improving video understanding and generation with
  better captions.
\newblock \emph{arXiv preprint arXiv:2406.04325}, 2024{\natexlab{a}}.

\bibitem[Chen et~al.(2024{\natexlab{b}})Chen, Xue, Li, Hu, Zhu, Li, Fang, Tang,
  Yang, Liu, He, Yin, Molchanov, Kautz, Fan, Zhu, Lu, and
  Han]{chen2024longvilascalinglongcontextvisual}
Chen, Y., Xue, F., Li, D., Hu, Q., Zhu, L., Li, X., Fang, Y., Tang, H., Yang,
  S., Liu, Z., He, Y., Yin, H., Molchanov, P., Kautz, J., Fan, L., Zhu, Y., Lu,
  Y., and Han, S.
\newblock Longvila: Scaling long-context visual language models for long
  videos.
\newblock 2024{\natexlab{b}}.

\bibitem[Ding et~al.(2024{\natexlab{a}})Ding, Qian, Dong, Zhang, Zang, Cao,
  Guo, Lin, and Wang]{ding2024sam2longenhancingsam2}
Ding, S., Qian, R., Dong, X., Zhang, P., Zang, Y., Cao, Y., Guo, Y., Lin, D.,
  and Wang, J.
\newblock Sam2long: Enhancing sam 2 for long video segmentation with a
  training-free memory tree.
\newblock \emph{arXiv preprint arXiv:2410.16268}, 2024{\natexlab{a}}.

\bibitem[Ding et~al.(2025)Ding, Wu, Zhao, Zang, Duan, Dong, Zhang, Cao, Lin,
  and Wang]{ding2025mm}
Ding, S., Wu, S., Zhao, X., Zang, Y., Duan, H., Dong, X., Zhang, P., Cao, Y.,
  Lin, D., and Wang, J.
\newblock Mm-ifengine: Towards multimodal instruction following.
\newblock \emph{arXiv preprint arXiv:2504.07957}, 2025.

\bibitem[Ding et~al.(2024{\natexlab{b}})Ding, Zhang, Zhang, Xu, Shang, Xu,
  Yang, and Yang]{ding2024longrope}
Ding, Y., Zhang, L.~L., Zhang, C., Xu, Y., Shang, N., Xu, J., Yang, F., and
  Yang, M.
\newblock {LongRoPE}: Extending llm context window beyond 2 million tokens.
\newblock \emph{arXiv preprint arXiv:2402.13753}, 2024{\natexlab{b}}.

\bibitem[Dong et~al.(2024)Dong, Zhang, Zang, Cao, Wang, Ouyang, Wei, Zhang,
  Duan, Cao, Zhang, Li, Yan, Gao, Zhang, Li, Li, Chen, He, Zhang, Qiao, Lin,
  and Wang]{internlmxcomposer2}
Dong, X., Zhang, P., Zang, Y., Cao, Y., Wang, B., Ouyang, L., Wei, X., Zhang,
  S., Duan, H., Cao, M., Zhang, W., Li, Y., Yan, H., Gao, Y., Zhang, X., Li,
  W., Li, J., Chen, K., He, C., Zhang, X., Qiao, Y., Lin, D., and Wang, J.
\newblock Internlm-xcomposer2: Mastering free-form text-image composition and
  comprehension in vision-language large model.
\newblock \emph{arXiv preprint arXiv:2401.16420}, 2024.

\bibitem[Dubey et~al.(2024)Dubey, Jauhri, Pandey, Kadian, Al-Dahle, Letman,
  Mathur, Schelten, Yang, Fan, et~al.]{dubey2024llama}
Dubey, A., Jauhri, A., Pandey, A., Kadian, A., Al-Dahle, A., Letman, A.,
  Mathur, A., Schelten, A., Yang, A., Fan, A., et~al.
\newblock The llama 3 herd of models.
\newblock \emph{arXiv preprint arXiv:2407.21783}, 2024.

\bibitem[Fabian Caba~Heilbron \& Niebles(2015)Fabian Caba~Heilbron and
  Niebles]{caba2015activitynet}
Fabian Caba~Heilbron, Victor~Escorcia, B.~G. and Niebles, J.~C.
\newblock {ActivityNet}: A large-scale video benchmark for human activity
  understanding.
\newblock In \emph{CVPR}, 2015.

\bibitem[Fu et~al.(2024)Fu, Dai, Luo, Li, Ren, Zhang, Wang, Zhou, Shen, Zhang,
  et~al.]{fu2024video}
Fu, C., Dai, Y., Luo, Y., Li, L., Ren, S., Zhang, R., Wang, Z., Zhou, C., Shen,
  Y., Zhang, M., et~al.
\newblock {Video-MME}: The first-ever comprehensive evaluation benchmark of
  multi-modal llms in video analysis.
\newblock \emph{arXiv preprint arXiv:2405.21075}, 2024.

\bibitem[Gao et~al.(2024)Gao, Liu, Li, Xie, Liu, Zhao, Chen, and
  Xiong]{gao2024tc}
Gao, M., Liu, J., Li, M., Xie, J., Liu, Q., Zhao, B., Chen, X., and Xiong, H.
\newblock {TC-LLaVA}: Rethinking the transfer from image to video understanding
  with temporal considerations.
\newblock \emph{arXiv preprint arXiv:2409.03206}, 2024.

\bibitem[Google(2025)]{googleimagesearch}
Google.
\newblock Google image search, 2025.
\newblock URL \url{https://images.google.com}.
\newblock Accessed: 2025-01-12.

\bibitem[Han et~al.(2024)Han, Wang, Peng, Xiong, Chen, Ji, and Wang]{han2024lm}
Han, C., Wang, Q., Peng, H., Xiong, W., Chen, Y., Ji, H., and Wang, S.
\newblock Lm-infinite: Zero-shot extreme length generalization for large
  language models.
\newblock In \emph{Proceedings of the 2024 Conference of the North American
  Chapter of the Association for Computational Linguistics: Human Language
  Technologies (Volume 1: Long Papers)}, pp.\  3991--4008, 2024.

\bibitem[Hsieh et~al.(2024)Hsieh, Sun, Kriman, Acharya, Rekesh, Jia, Zhang, and
  Ginsburg]{hsieh2024ruler}
Hsieh, C.-P., Sun, S., Kriman, S., Acharya, S., Rekesh, D., Jia, F., Zhang, Y.,
  and Ginsburg, B.
\newblock Ruler: What's the real context size of your long-context language
  models?
\newblock \emph{arXiv preprint arXiv:2404.06654}, 2024.

\bibitem[Huang et~al.(2023)Huang, Wang, Chen, Song, and
  Zhu]{huang2023vtimellmempowerllmgrasp}
Huang, B., Wang, X., Chen, H., Song, Z., and Zhu, W.
\newblock Vtimellm: Empower llm to grasp video moments, 2023.
\newblock URL \url{https://arxiv.org/abs/2311.18445}.

\bibitem[Huang et~al.(2024)Huang, Dong, Zhang, Wang, He, Wang, Lin, Zhang, and
  Yu]{huang2024operaalleviatinghallucinationmultimodal}
Huang, Q., Dong, X., Zhang, P., Wang, B., He, C., Wang, J., Lin, D., Zhang, W.,
  and Yu, N.
\newblock Opera: Alleviating hallucination in multi-modal large language models
  via over-trust penalty and retrospection-allocation, 2024.
\newblock URL \url{https://arxiv.org/abs/2311.17911}.

\bibitem[Jin et~al.(2023)Jin, Takanobu, Zhang, Cao, and Yuan]{jin2023chatunivi}
Jin, P., Takanobu, R., Zhang, C., Cao, X., and Yuan, L.
\newblock Chat-univi: Unified visual representation empowers large language
  models with image and video understanding.
\newblock \emph{arXiv preprint arXiv:2311.08046}, 2023.

\bibitem[Kamradt(2023)]{niah}
Kamradt, G.
\newblock Needle in a haystack - pressure testing llms.
\newblock \url{https://github.com/gkamradt/LLMTest_NeedleInAHaystack}, 2023.

\bibitem[Kay et~al.(2017)Kay, Carreira, Simonyan, Zhang, Hillier,
  Vijayanarasimhan, Viola, Green, Back, Natsev, et~al.]{kay2017kinetics}
Kay, W., Carreira, J., Simonyan, K., Zhang, B., Hillier, C., Vijayanarasimhan,
  S., Viola, F., Green, T., Back, T., Natsev, P., et~al.
\newblock The kinetics human action video dataset.
\newblock \emph{arXiv preprint arXiv:1705.06950}, 2017.

\bibitem[Kwon et~al.(2023)Kwon, Li, Zhuang, Sheng, Zheng, Yu, Gonzalez, Zhang,
  and Stoica]{kwon2023efficient}
Kwon, W., Li, Z., Zhuang, S., Sheng, Y., Zheng, L., Yu, C.~H., Gonzalez, J.~E.,
  Zhang, H., and Stoica, I.
\newblock Efficient memory management for large language model serving with
  pagedattention.
\newblock In \emph{ACM SIGOPS}, 2023.

\bibitem[Labs(2023)]{flux2023}
Labs, B.~F.
\newblock Flux.
\newblock \url{https://github.com/black-forest-labs/flux}, 2023.

\bibitem[LangChain(2024)]{multi_niah}
LangChain.
\newblock Multi needle in a haystack.
\newblock \url{hhttps://blog.langchain.dev/multi-needle-in-a-haystack/}, 2024.

\bibitem[Lei et~al.(2021)Lei, Li, Zhou, Gan, Berg, Bansal, and
  Liu]{lei2021moreclipbertvideoandlanguagelearning}
Lei, J., Li, L., Zhou, L., Gan, Z., Berg, T.~L., Bansal, M., and Liu, J.
\newblock Less is more: Clipbert for video-and-language learning via sparse
  sampling, 2021.
\newblock URL \url{https://arxiv.org/abs/2102.06183}.

\bibitem[Li et~al.(2023)Li, He, Wang, Li, Wang, Luo, Wang, Wang, and
  Qiao]{2023videochat}
Li, K., He, Y., Wang, Y., Li, Y., Wang, W., Luo, P., Wang, Y., Wang, L., and
  Qiao, Y.
\newblock {VideoChat}: Chat-centric video understanding.
\newblock \emph{arXiv preprint arXiv:2305.06355}, 2023.

\bibitem[Li et~al.(2024{\natexlab{a}})Li, Liu, Yao, Zhang, An, Wang, Sun, Kong,
  and Liu]{li2024temporal}
Li, L., Liu, Y., Yao, L., Zhang, P., An, C., Wang, L., Sun, X., Kong, L., and
  Liu, Q.
\newblock Temporal reasoning transfer from text to video.
\newblock \emph{arXiv preprint arXiv:2410.06166}, 2024{\natexlab{a}}.

\bibitem[Li et~al.(2024{\natexlab{b}})Li, Wang, and Jia]{li2024llamavid}
Li, Y., Wang, C., and Jia, J.
\newblock Llama-vid: An image is worth 2 tokens in large language models.
\newblock 2024{\natexlab{b}}.

\bibitem[Li et~al.(2025)Li, Niu, Miao, Ge, Zhou, He, Dong, Duan, Ding, Qian,
  Zhang, Zang, Cao, He, and Wang]{li2025ovobenchfarvideollmsrealworld}
Li, Y., Niu, J., Miao, Z., Ge, C., Zhou, Y., He, Q., Dong, X., Duan, H., Ding,
  S., Qian, R., Zhang, P., Zang, Y., Cao, Y., He, C., and Wang, J.
\newblock Ovo-bench: How far is your video-llms from real-world online video
  understanding?, 2025.
\newblock URL \url{https://arxiv.org/abs/2501.05510}.

\bibitem[Lin et~al.(2023{\natexlab{a}})Lin, Zhu, Ye, Ning, Jin, and
  Yuan]{lin2023video}
Lin, B., Zhu, B., Ye, Y., Ning, M., Jin, P., and Yuan, L.
\newblock Video-llava: Learning united visual representation by alignment
  before projection.
\newblock \emph{arXiv preprint arXiv:2311.10122}, 2023{\natexlab{a}}.

\bibitem[Lin et~al.(2023{\natexlab{b}})Lin, Zhu, Ye, Ning, Jin, and
  Yuan]{lin2023videollava}
Lin, B., Zhu, B., Ye, Y., Ning, M., Jin, P., and Yuan, L.
\newblock Video-llava: Learning united visual representation by alignment
  before projection.
\newblock \emph{arXiv preprint arXiv:2311.10122}, 2023{\natexlab{b}}.

\bibitem[Lin et~al.(2023{\natexlab{c}})Lin, Ahmed, Li, Lin, Azarnasab, Yang,
  Wang, Liang, Liu, Lu, Liu, and Wang]{lin2023mmvidadvancingvideounderstanding}
Lin, K., Ahmed, F., Li, L., Lin, C.-C., Azarnasab, E., Yang, Z., Wang, J.,
  Liang, L., Liu, Z., Lu, Y., Liu, C., and Wang, L.
\newblock Mm-vid: Advancing video understanding with gpt-4v(ision),
  2023{\natexlab{c}}.
\newblock URL \url{https://arxiv.org/abs/2310.19773}.

\bibitem[Liu et~al.(2023{\natexlab{a}})Liu, Li, Wu, and Lee]{liu2023llava}
Liu, H., Li, C., Wu, Q., and Lee, Y.~J.
\newblock Visual instruction tuning.
\newblock In \emph{NeurIPS}, 2023{\natexlab{a}}.

\bibitem[Liu et~al.(2023{\natexlab{b}})Liu, Yan, Zhang, An, Qiu, and
  Lin]{liu2023scaling}
Liu, X., Yan, H., Zhang, S., An, C., Qiu, X., and Lin, D.
\newblock Scaling laws of rope-based extrapolation.
\newblock \emph{arXiv preprint arXiv:2310.05209}, 2023{\natexlab{b}}.

\bibitem[Liu et~al.(2024{\natexlab{a}})Liu, Ma, Qi, Wu, Shan, and
  Chen]{liu2024etbenchopenendedeventlevel}
Liu, Y., Ma, Z., Qi, Z., Wu, Y., Shan, Y., and Chen, C.~W.
\newblock E.t. bench: Towards open-ended event-level video-language
  understanding, 2024{\natexlab{a}}.
\newblock URL \url{https://arxiv.org/abs/2409.18111}.

\bibitem[Liu et~al.(2024{\natexlab{b}})Liu, Chu, Zang, Wei, Dong, Zhang, Liang,
  Xiong, Qiao, Lin, et~al.]{liu2024mmdu}
Liu, Z., Chu, T., Zang, Y., Wei, X., Dong, X., Zhang, P., Liang, Z., Xiong, Y.,
  Qiao, Y., Lin, D., et~al.
\newblock Mmdu: A multi-turn multi-image dialog understanding benchmark and
  instruction-tuning dataset for lvlms.
\newblock \emph{arXiv preprint arXiv:2406.11833}, 2024{\natexlab{b}}.

\bibitem[Liu et~al.(2024{\natexlab{c}})Liu, Sun, Zang, Li, Zhang, Dong, Xiong,
  Lin, and Wang]{liu2024rar}
Liu, Z., Sun, Z., Zang, Y., Li, W., Zhang, P., Dong, X., Xiong, Y., Lin, D.,
  and Wang, J.
\newblock Rar: Retrieving and ranking augmented mllms for visual recognition,
  2024{\natexlab{c}}.

\bibitem[Liu et~al.(2024{\natexlab{d}})Liu, Zang, Dong, Zhang, Cao, Duan, He,
  Xiong, Lin, and Wang]{liu2024mia}
Liu, Z., Zang, Y., Dong, X., Zhang, P., Cao, Y., Duan, H., He, C., Xiong, Y.,
  Lin, D., and Wang, J.
\newblock Mia-dpo: Multi-image augmented direct preference optimization for
  large vision-language models.
\newblock \emph{arXiv preprint arXiv:2410.17637}, 2024{\natexlab{d}}.

\bibitem[Luo et~al.(2023)Luo, Zhao, Yang, Dong, Li, Lu, Wang, Hu, Qiu, and
  Wei]{luo2023valleyvideoassistantlarge}
Luo, R., Zhao, Z., Yang, M., Dong, J., Li, D., Lu, P., Wang, T., Hu, L., Qiu,
  M., and Wei, Z.
\newblock Valley: Video assistant with large language model enhanced ability,
  2023.
\newblock URL \url{https://arxiv.org/abs/2306.07207}.

\bibitem[Maaz et~al.(2024{\natexlab{a}})Maaz, Rasheed, Khan, and
  Khan]{Maaz2023VideoChatGPT}
Maaz, M., Rasheed, H., Khan, S., and Khan, F.~S.
\newblock Video-chatgpt: Towards detailed video understanding via large vision
  and language models.
\newblock In \emph{Proceedings of the 62nd Annual Meeting of the Association
  for Computational Linguistics (ACL 2024)}, 2024{\natexlab{a}}.

\bibitem[Maaz et~al.(2024{\natexlab{b}})Maaz, Rasheed, Khan, and
  Khan]{maaz2024videochatgptdetailedvideounderstanding}
Maaz, M., Rasheed, H., Khan, S., and Khan, F.~S.
\newblock Video-chatgpt: Towards detailed video understanding via large vision
  and language models.
\newblock In \emph{Proceedings of the 62nd Annual Meeting of the Association
  for Computational Linguistics (ACL 2024)}, 2024{\natexlab{b}}.

\bibitem[Men et~al.(2024)Men, Xu, Wang, Zhang, Lin, Han, and Chen]{men2024base}
Men, X., Xu, M., Wang, B., Zhang, Q., Lin, H., Han, X., and Chen, W.
\newblock Base of rope bounds context length.
\newblock \emph{arXiv preprint arXiv:2405.14591}, 2024.

\bibitem[Peng et~al.(2023)Peng, Quesnelle, Fan, and Shippole]{peng2023yarn}
Peng, B., Quesnelle, J., Fan, H., and Shippole, E.
\newblock Yarn: Efficient context window extension of large language models.
\newblock \emph{arXiv preprint arXiv:2309.00071}, 2023.

\bibitem[Qian et~al.(2025{\natexlab{a}})Qian, Ding, Dong, Zhang, Zang, Cao,
  Lin, and Wang]{qian2025dispiderenablingvideollms}
Qian, R., Ding, S., Dong, X., Zhang, P., Zang, Y., Cao, Y., Lin, D., and Wang,
  J.
\newblock Dispider: Enabling video llms with active real-time interaction via
  disentangled perception, decision, and reaction.
\newblock \emph{arXiv preprint arXiv:2501.03218}, 2025{\natexlab{a}}.

\bibitem[Qian et~al.(2025{\natexlab{b}})Qian, Dong, Zhang, Zang, Ding, Lin, and
  Wang]{qian2024streaminglongvideounderstanding}
Qian, R., Dong, X., Zhang, P., Zang, Y., Ding, S., Lin, D., and Wang, J.
\newblock Streaming long video understanding with large language models.
\newblock \emph{Advances in Neural Information Processing Systems},
  37:\penalty0 119336--119360, 2025{\natexlab{b}}.

\bibitem[Radford et~al.(2021)Radford, Kim, Hallacy, Ramesh, Goh, Agarwal,
  Sastry, Askell, Mishkin, Clark, Krueger, and
  Sutskever]{radford2021learningtransferablevisualmodels}
Radford, A., Kim, J.~W., Hallacy, C., Ramesh, A., Goh, G., Agarwal, S., Sastry,
  G., Askell, A., Mishkin, P., Clark, J., Krueger, G., and Sutskever, I.
\newblock Learning transferable visual models from natural language
  supervision, 2021.
\newblock URL \url{https://arxiv.org/abs/2103.00020}.

\bibitem[Su(2024{\natexlab{a}})]{kexuefm10040}
Su, J.
\newblock Transformer upgrade path: 17. insights into multimodal positional
  encoding, March 2024{\natexlab{a}}.
\newblock URL \url{https://spaces.ac.cn/archives/10040}.

\bibitem[Su(2024{\natexlab{b}})]{kexuefm10352}
Su, J.
\newblock A brief discussion on multimodal thinking: 3. positional encoding,
  Sep 2024{\natexlab{b}}.
\newblock URL \url{https://spaces.ac.cn/archives/10352}.

\bibitem[Su et~al.(2024)Su, Ahmed, Lu, Pan, Bo, and Liu]{su2024roformer}
Su, J., Ahmed, M., Lu, Y., Pan, S., Bo, W., and Liu, Y.
\newblock {RoFormer}: Enhanced transformer with rotary position embedding.
\newblock \emph{Neurocomputing}, 2024.

\bibitem[Sun et~al.(2024)Sun, Zhang, He, Li, Cheng, Liu, Yan, Shao, Tang,
  Zhang, Zhao, Chen, Zheng, Zhou, Li, Zhan, Zhou, Li, Yang, Wu, Yin, Huang,
  Jiang, and Qiu]{Sun2024MOSS}
Sun, T., Zhang, X., He, Z., Li, P., Cheng, Q., Liu, X., Yan, H., Shao, Y.,
  Tang, Q., Zhang, S., Zhao, X., Chen, K., Zheng, Y., Zhou, Z., Li, R., Zhan,
  J., Zhou, Y., Li, L., Yang, X., Wu, L., Yin, Z., Huang, X., Jiang, Y.-G., and
  Qiu, X.
\newblock Moss: An open conversational large language model.
\newblock \emph{Machine Intelligence Research}, 2024.
\newblock ISSN 2731-5398.
\newblock \doi{10.1007/s11633-024-1502-8}.
\newblock URL \url{https://github.com/OpenMOSS/MOSS}.

\bibitem[Sun et~al.(2023)Sun, Fang, Wu, Zhang, Zang, Kong, Xiong, Lin, and
  Wang]{sun2023alphaclip}
Sun, Z., Fang, Y., Wu, T., Zhang, P., Zang, Y., Kong, S., Xiong, Y., Lin, D.,
  and Wang, J.
\newblock Alpha-clip: A clip model focusing on wherever you want, 2023.

\bibitem[Team et~al.(2024)Team, Mesnard, Hardin, Dadashi, Bhupatiraju, Pathak,
  Sifre, Rivière, Kale, Love, Tafti, Hussenot, Sessa, Chowdhery, Roberts,
  Barua, Botev, Castro-Ros, Slone, Héliou, Tacchetti, Bulanova, Paterson,
  Tsai, Shahriari, Lan, Choquette-Choo, Crepy, Cer, Ippolito, Reid,
  Buchatskaya, Ni, Noland, Yan, Tucker, Muraru, Rozhdestvenskiy, Michalewski,
  Tenney, Grishchenko, Austin, Keeling, Labanowski, Lespiau, Stanway, Brennan,
  Chen, Ferret, Chiu, Mao-Jones, Lee, Yu, Millican, Sjoesund, Lee, Dixon, Reid,
  Mikuła, Wirth, Sharman, Chinaev, Thain, Bachem, Chang, Wahltinez, Bailey,
  Michel, Yotov, Chaabouni, Comanescu, Jana, Anil, McIlroy, Liu, Mullins,
  Smith, Borgeaud, Girgin, Douglas, Pandya, Shakeri, De, Klimenko, Hennigan,
  Feinberg, Stokowiec, hui Chen, Ahmed, Gong, Warkentin, Peran, Giang, Farabet,
  Vinyals, Dean, Kavukcuoglu, Hassabis, Ghahramani, Eck, Barral, Pereira,
  Collins, Joulin, Fiedel, Senter, Andreev, and
  Kenealy]{gemmateam2024gemmaopenmodelsbased}
Team, G., Mesnard, T., Hardin, C., Dadashi, R., Bhupatiraju, S., Pathak, S.,
  Sifre, L., Rivière, M., Kale, M.~S., Love, J., Tafti, P., Hussenot, L.,
  Sessa, P.~G., Chowdhery, A., Roberts, A., Barua, A., Botev, A., Castro-Ros,
  A., Slone, A., Héliou, A., Tacchetti, A., Bulanova, A., Paterson, A., Tsai,
  B., Shahriari, B., Lan, C.~L., Choquette-Choo, C.~A., Crepy, C., Cer, D.,
  Ippolito, D., Reid, D., Buchatskaya, E., Ni, E., Noland, E., Yan, G., Tucker,
  G., Muraru, G.-C., Rozhdestvenskiy, G., Michalewski, H., Tenney, I.,
  Grishchenko, I., Austin, J., Keeling, J., Labanowski, J., Lespiau, J.-B.,
  Stanway, J., Brennan, J., Chen, J., Ferret, J., Chiu, J., Mao-Jones, J., Lee,
  K., Yu, K., Millican, K., Sjoesund, L.~L., Lee, L., Dixon, L., Reid, M.,
  Mikuła, M., Wirth, M., Sharman, M., Chinaev, N., Thain, N., Bachem, O.,
  Chang, O., Wahltinez, O., Bailey, P., Michel, P., Yotov, P., Chaabouni, R.,
  Comanescu, R., Jana, R., Anil, R., McIlroy, R., Liu, R., Mullins, R., Smith,
  S.~L., Borgeaud, S., Girgin, S., Douglas, S., Pandya, S., Shakeri, S., De,
  S., Klimenko, T., Hennigan, T., Feinberg, V., Stokowiec, W., hui Chen, Y.,
  Ahmed, Z., Gong, Z., Warkentin, T., Peran, L., Giang, M., Farabet, C.,
  Vinyals, O., Dean, J., Kavukcuoglu, K., Hassabis, D., Ghahramani, Z., Eck,
  D., Barral, J., Pereira, F., Collins, E., Joulin, A., Fiedel, N., Senter, E.,
  Andreev, A., and Kenealy, K.
\newblock Gemma: Open models based on gemini research and technology, 2024.
\newblock URL \url{https://arxiv.org/abs/2403.08295}.

\bibitem[Touvron et~al.(2023{\natexlab{a}})Touvron, Lavril, Izacard, Martinet,
  Lachaux, Lacroix, Rozière, Goyal, Hambro, Azhar, Rodriguez, Joulin, Grave,
  and Lample]{touvron2023llamaopenefficientfoundation}
Touvron, H., Lavril, T., Izacard, G., Martinet, X., Lachaux, M.-A., Lacroix,
  T., Rozière, B., Goyal, N., Hambro, E., Azhar, F., Rodriguez, A., Joulin,
  A., Grave, E., and Lample, G.
\newblock Llama: Open and efficient foundation language models,
  2023{\natexlab{a}}.
\newblock URL \url{https://arxiv.org/abs/2302.13971}.

\bibitem[Touvron et~al.(2023{\natexlab{b}})Touvron, Martin, Stone, Albert,
  Almahairi, Babaei, Bashlykov, Batra, Bhargava, Bhosale,
  et~al.]{touvron2023llama}
Touvron, H., Martin, L., Stone, K., Albert, P., Almahairi, A., Babaei, Y.,
  Bashlykov, N., Batra, S., Bhargava, P., Bhosale, S., et~al.
\newblock {LLaMA 2}: Open foundation and fine-tuned chat models.
\newblock \emph{arXiv preprint arXiv:2307.09288}, 2023{\natexlab{b}}.

\bibitem[Wang et~al.(2025)Wang, Shi, Tan, Qin, Wang, Zhang, Nambi, Ganu, and
  Wang]{wang2024multimodal}
Wang, H., Shi, H., Tan, S., Qin, W., Wang, W., Zhang, T., Nambi, A., Ganu, T.,
  and Wang, H.
\newblock Multimodal needle in a haystack: Benchmarking long-context capability
  of multimodal large language models.
\newblock In \emph{Proceedings of the 2025 Conference of the North American
  Chapter of the Association for Computational Linguistics}, 2025.

\bibitem[Wang et~al.(2023)Wang, Chen, Luo, Dai, Yuan, Wu, and
  Jiang]{wang2023chatvideotrackletcentricmultimodalversatile}
Wang, J., Chen, D., Luo, C., Dai, X., Yuan, L., Wu, Z., and Jiang, Y.-G.
\newblock Chatvideo: A tracklet-centric multimodal and versatile video
  understanding system, 2023.
\newblock URL \url{https://arxiv.org/abs/2304.14407}.

\bibitem[Wang et~al.(2024{\natexlab{a}})Wang, Bai, Tan, Wang, Fan, Bai, Chen,
  Liu, Wang, Ge, et~al.]{wang2024qwen2}
Wang, P., Bai, S., Tan, S., Wang, S., Fan, Z., Bai, J., Chen, K., Liu, X.,
  Wang, J., Ge, W., et~al.
\newblock {Qwen2-VL}: Enhancing vision-language model's perception of the world
  at any resolution.
\newblock \emph{arXiv preprint arXiv:2409.12191}, 2024{\natexlab{a}}.

\bibitem[Wang et~al.(2024{\natexlab{b}})Wang, Zhang, Ren, Duan, Li, Liu, Hu,
  Chen, Zhang, Lu, et~al.]{wang2024needle}
Wang, W., Zhang, S., Ren, Y., Duan, Y., Li, T., Liu, S., Hu, M., Chen, Z.,
  Zhang, K., Lu, L., et~al.
\newblock Needle in a multimodal haystack.
\newblock \emph{arXiv preprint arXiv:2406.07230}, 2024{\natexlab{b}}.

\bibitem[Wang et~al.(2024{\natexlab{c}})Wang, Song, Chen, Zhang, and
  Wang]{wang2024longllavascalingmultimodalllms}
Wang, X., Song, D., Chen, S., Zhang, C., and Wang, B.
\newblock Longllava: Scaling multi-modal llms to 1000 images efficiently via a
  hybrid architecture, 2024{\natexlab{c}}.
\newblock URL \url{https://arxiv.org/abs/2409.02889}.

\bibitem[Wang et~al.(2022)Wang, Li, Li, He, Huang, Zhao, Zhang, Xu, Liu, Wang,
  Xing, Chen, Pan, Yu, Wang, Wang, and
  Qiao]{wang2022internvideogeneralvideofoundation}
Wang, Y., Li, K., Li, Y., He, Y., Huang, B., Zhao, Z., Zhang, H., Xu, J., Liu,
  Y., Wang, Z., Xing, S., Chen, G., Pan, J., Yu, J., Wang, Y., Wang, L., and
  Qiao, Y.
\newblock Internvideo: General video foundation models via generative and
  discriminative learning, 2022.
\newblock URL \url{https://arxiv.org/abs/2212.03191}.

\bibitem[Wang et~al.(2024{\natexlab{d}})Wang, Wang, Zhao, Xie, and
  Zheng]{videohallucer}
Wang, Y., Wang, Y., Zhao, D., Xie, C., and Zheng, Z.
\newblock Videohallucer: Evaluating intrinsic and extrinsic hallucinations in
  large video-language models.
\newblock \emph{arxiv}, 2024{\natexlab{d}}.

\bibitem[Wang et~al.(2024{\natexlab{e}})Wang, Yu, Stengel-Eskin, Yoon, Cheng,
  Bertasius, and Bansal]{wang2024videotreeadaptivetreebasedvideo}
Wang, Z., Yu, S., Stengel-Eskin, E., Yoon, J., Cheng, F., Bertasius, G., and
  Bansal, M.
\newblock Videotree: Adaptive tree-based video representation for llm reasoning
  on long videos.
\newblock \emph{arXiv preprint arXiv:2405.19209}, 2024{\natexlab{e}}.

\bibitem[Wu et~al.(2024{\natexlab{a}})Wu, Li, Chen, and
  Li]{wu2024longvideobench}
Wu, H., Li, D., Chen, B., and Li, J.
\newblock {LongVideoBench}: A benchmark for long-context interleaved
  video-language understanding.
\newblock \emph{arXiv preprint arXiv:2407.15754}, 2024{\natexlab{a}}.

\bibitem[Wu et~al.(2024{\natexlab{b}})Wu, Biamby, , Quenum, Gupta, Gonzalez,
  Darrell, and Chan]{wu2024visual}
Wu, T.-H., Biamby, G., , Quenum, J., Gupta, R., Gonzalez, J.~E., Darrell, T.,
  and Chan, D.~M.
\newblock Visual haystacks: A vision-centric needle-in-a-haystack benchmark.
\newblock \emph{arXiv preprint arXiv:2407.13766}, 2024{\natexlab{b}}.

\bibitem[Xiao et~al.(2023)Xiao, Tian, Chen, Han, and Lewis]{xiao2023efficient}
Xiao, G., Tian, Y., Chen, B., Han, S., and Lewis, M.
\newblock Efficient streaming language models with attention sinks.
\newblock \emph{arXiv preprint arXiv:2309.17453}, 2023.

\bibitem[Xing et~al.(2024)Xing, Huang, Dong, Lu, Zhang, Zang, Cao, He, Wang,
  Wu, and Lin]{xing2024pyramiddropacceleratinglargevisionlanguage}
Xing, L., Huang, Q., Dong, X., Lu, J., Zhang, P., Zang, Y., Cao, Y., He, C.,
  Wang, J., Wu, F., and Lin, D.
\newblock Pyramiddrop: Accelerating your large vision-language models via
  pyramid visual redundancy reduction, 2024.
\newblock URL \url{https://arxiv.org/abs/2410.17247}.

\bibitem[Xu et~al.(2021)Xu, Ghosh, Huang, Okhonko, Aghajanyan, Metze,
  Zettlemoyer, and
  Feichtenhofer]{xu2021videoclipcontrastivepretrainingzeroshot}
Xu, H., Ghosh, G., Huang, P.-Y., Okhonko, D., Aghajanyan, A., Metze, F.,
  Zettlemoyer, L., and Feichtenhofer, C.
\newblock Videoclip: Contrastive pre-training for zero-shot video-text
  understanding, 2021.
\newblock URL \url{https://arxiv.org/abs/2109.14084}.

\bibitem[Xu et~al.(2024)Xu, Zhao, Zhou, Lin, Ng, and
  Feng]{xu2024pllavaparameterfreellava}
Xu, L., Zhao, Y., Zhou, D., Lin, Z., Ng, S.~K., and Feng, J.
\newblock Pllava : Parameter-free llava extension from images to videos for
  video dense captioning, 2024.
\newblock URL \url{https://arxiv.org/abs/2404.16994}.

\bibitem[Xue et~al.(2022)Xue, Hang, Zeng, Sun, Liu, Yang, Fu, and
  Guo]{xue2022hdvila}
Xue, H., Hang, T., Zeng, Y., Sun, Y., Liu, B., Yang, H., Fu, J., and Guo, B.
\newblock Advancing high-resolution video-language representation with
  large-scale video transcriptions.
\newblock In \emph{CVPR}, 2022.

\bibitem[Yang et~al.(2024{\natexlab{a}})Yang, Yang, Hui, Zheng, Yu, Zhou, Li,
  Li, Liu, Huang, et~al.]{yang2024qwen2}
Yang, A., Yang, B., Hui, B., Zheng, B., Yu, B., Zhou, C., Li, C., Li, C., Liu,
  D., Huang, F., et~al.
\newblock Qwen2 technical report.
\newblock \emph{arXiv preprint arXiv:2407.10671}, 2024{\natexlab{a}}.

\bibitem[Yang et~al.(2024{\natexlab{b}})Yang, Yang, Zhang, Hui, Zheng, Yu, Li,
  Liu, Huang, Wei, et~al.]{yang2024qwen25}
Yang, A., Yang, B., Zhang, B., Hui, B., Zheng, B., Yu, B., Li, C., Liu, D.,
  Huang, F., Wei, H., et~al.
\newblock Qwen2. 5 technical report.
\newblock \emph{arXiv preprint arXiv:2412.15115}, 2024{\natexlab{b}}.

\bibitem[Yuan et~al.(2024)Yuan, Ning, Zhou, Yang, Li, Zhuang, Tan, Yao, Lin,
  Li, et~al.]{yuan2024lv}
Yuan, T., Ning, X., Zhou, D., Yang, Z., Li, S., Zhuang, M., Tan, Z., Yao, Z.,
  Lin, D., Li, B., et~al.
\newblock Lv-eval: A balanced long-context benchmark with 5 length levels up to
  256k.
\newblock \emph{arXiv preprint arXiv:2402.05136}, 2024.

\bibitem[Zang et~al.(2025)Zang, Dong, Zhang, Cao, Liu, Ding, Wu, Ma, Duan,
  Zhang, et~al.]{internlmxcomposer2_5_reward}
Zang, Y., Dong, X., Zhang, P., Cao, Y., Liu, Z., Ding, S., Wu, S., Ma, Y.,
  Duan, H., Zhang, W., et~al.
\newblock Internlm-xcomposer2. 5-reward: A simple yet effective multi-modal
  reward model.
\newblock \emph{arXiv preprint arXiv:2501.12368}, 2025.

\bibitem[Zhang et~al.(2024{\natexlab{a}})Zhang, Zhang, Dong, Zang, and
  Wang]{zhang2024longclip}
Zhang, B., Zhang, P., Dong, X., Zang, Y., and Wang, J.
\newblock Long-clip: Unlocking the long-text capability of clip.
\newblock \emph{arXiv preprint arXiv:2403.15378}, 2024{\natexlab{a}}.

\bibitem[Zhang et~al.(2023{\natexlab{a}})Zhang, Li, and
  Bing]{zhang2023videollamainstructiontunedaudiovisuallanguage}
Zhang, H., Li, X., and Bing, L.
\newblock Video-llama: An instruction-tuned audio-visual language model for
  video understanding, 2023{\natexlab{a}}.
\newblock URL \url{https://arxiv.org/abs/2306.02858}.

\bibitem[Zhang et~al.(2023{\natexlab{b}})Zhang, Dong, Wang, Cao, Xu, Ouyang,
  Zhao, Ding, Zhang, Duan, Zhang, Yan, Zhang, Li, Li, Chen, He, Zhang, Qiao,
  Lin, and Wang]{internlmxcomposer}
Zhang, P., Dong, X., Wang, B., Cao, Y., Xu, C., Ouyang, L., Zhao, Z., Ding, S.,
  Zhang, S., Duan, H., Zhang, W., Yan, H., Zhang, X., Li, W., Li, J., Chen, K.,
  He, C., Zhang, X., Qiao, Y., Lin, D., and Wang, J.
\newblock Internlm-xcomposer: A vision-language large model for advanced
  text-image comprehension and composition.
\newblock \emph{arXiv preprint arXiv:2309.15112}, 2023{\natexlab{b}}.

\bibitem[Zhang et~al.(2024{\natexlab{b}})Zhang, Dong, Cao, Zang, Qian, Wei,
  Chen, Li, Niu, Ding, Guo, Duan, Chen, Lv, Nie, Zhang, Wang, Zhang, Zhang, Ge,
  Li, Li, Tu, He, Zhang, Chen, Qiao, Lin, and Wang]{internlmxcomposer2_5_OL}
Zhang, P., Dong, X., Cao, Y., Zang, Y., Qian, R., Wei, X., Chen, L., Li, Y.,
  Niu, J., Ding, S., Guo, Q., Duan, H., Chen, X., Lv, H., Nie, Z., Zhang, M.,
  Wang, B., Zhang, W., Zhang, X., Ge, J., Li, W., Li, J., Tu, Z., He, C.,
  Zhang, X., Chen, K., Qiao, Y., Lin, D., and Wang, J.
\newblock Internlm-xcomposer2.5-omnilive: A comprehensive multimodal system for
  long-term streaming video and audio interactions.
\newblock \emph{arXiv preprint arXiv:2412.09596}, 2024{\natexlab{b}}.

\bibitem[Zhang et~al.(2024{\natexlab{c}})Zhang, Dong, Zang, Cao, Qian, Chen,
  Guo, Duan, Wang, Ouyang, Zhang, Zhang, Li, Gao, Sun, Zhang, Li, Li, Wang,
  Yan, He, Zhang, Chen, Dai, Qiao, Lin, and Wang]{internlmxcomposer2_5}
Zhang, P., Dong, X., Zang, Y., Cao, Y., Qian, R., Chen, L., Guo, Q., Duan, H.,
  Wang, B., Ouyang, L., Zhang, S., Zhang, W., Li, Y., Gao, Y., Sun, P., Zhang,
  X., Li, W., Li, J., Wang, W., Yan, H., He, C., Zhang, X., Chen, K., Dai, J.,
  Qiao, Y., Lin, D., and Wang, J.
\newblock Internlm-xcomposer-2.5: A versatile large vision language model
  supporting long-contextual input and output.
\newblock \emph{arXiv preprint arXiv:2407.03320}, 2024{\natexlab{c}}.

\bibitem[Zhang et~al.(2024{\natexlab{d}})Zhang, Zhang, Li, Zeng, Yang, Zhang,
  Wang, Tan, Li, and Liu]{zhang2024longva}
Zhang, P., Zhang, K., Li, B., Zeng, G., Yang, J., Zhang, Y., Wang, Z., Tan, H.,
  Li, C., and Liu, Z.
\newblock Long context transfer from language to vision.
\newblock \emph{arXiv preprint arXiv:2406.16852}, 2024{\natexlab{d}}.
\newblock URL \url{https://arxiv.org/abs/2406.16852}.

\bibitem[Zhang et~al.(2025)Zhang, Fang, Yang, and
  Feng]{zhang2025llavaminiefficientimagevideo}
Zhang, S., Fang, Q., Yang, Z., and Feng, Y.
\newblock Llava-mini: Efficient image and video large multimodal models with
  one vision token, 2025.
\newblock URL \url{https://arxiv.org/abs/2501.03895}.

\bibitem[Zhang et~al.(2024{\natexlab{e}})Zhang, Wu, Li, Li, Ma, Liu, and
  Li]{zhang2024video}
Zhang, Y., Wu, J., Li, W., Li, B., Ma, Z., Liu, Z., and Li, C.
\newblock Video instruction tuning with synthetic data.
\newblock \emph{arXiv preprint arXiv:2410.02713}, 2024{\natexlab{e}}.

\bibitem[Zhao et~al.(2024{\natexlab{a}})Zhao, Ding, Zhang, Huang, Cao, Wang,
  Wang, Fang, Wang, Zhai, Duan, Yang, and
  Chen]{zhao2025omnialignvenhancedalignmentmllms}
Zhao, X., Ding, S., Zhang, Z., Huang, H., Cao, M., Wang, W., Wang, J., Fang,
  X., Wang, W., Zhai, G., Duan, H., Yang, H., and Chen, K.
\newblock Omnialign-v: Towards enhanced alignment of mllms with human
  preference.
\newblock \emph{arXiv preprint arXiv:2502.18411}, 2024{\natexlab{a}}.

\bibitem[Zhao et~al.(2024{\natexlab{b}})Zhao, Lu, Huo, Du, Yue, Guo, Wang,
  Chen, and Liu]{zhao2024videoniah}
Zhao, Z., Lu, H., Huo, Y., Du, Y., Yue, T., Guo, L., Wang, B., Chen, W., and
  Liu, J.
\newblock Needle in a video haystack: A scalable synthetic framework for
  benchmarking video mllms.
\newblock \emph{arXiv preprint}, 2024{\natexlab{b}}.

\bibitem[Zhou et~al.(2024)Zhou, Shu, Zhao, Wu, Xiao, Yang, Xiong, Zhang, Huang,
  and Liu]{zhou2024mlvu}
Zhou, J., Shu, Y., Zhao, B., Wu, B., Xiao, S., Yang, X., Xiong, Y., Zhang, B.,
  Huang, T., and Liu, Z.
\newblock {MLVU}: A comprehensive benchmark for multi-task long video
  understanding.
\newblock \emph{arXiv preprint arXiv:2406.04264}, 2024.

\end{thebibliography}
\bibliographystyle{icml2025}

\newpage
\appendix
\onecolumn

\section*{\centering Appendix}

This appendix provides additional resources to further enhance the understanding of our work.
In Section \ref{app:MORE_EXPERIMENTS}, we present ablation studies and extrapolation experiments extending up to 128k, which are included here due to space constraints in the main text.
Section \ref{appendix:benchmarks} offers a more detailed discussion of the benchmarks used for evaluation.
Section \ref{app:related_work} reviews related work on video LLMs and video haystack retrieval.
Section \ref{app:video_niah_d} showcases examples from our proposed \textbf{V-NIAH-D} benchmark.
In Section \ref{app:attention_analysis}, we provide additional attention visualizations to further support the observations discussed in Figure \ref{fig:attention_analysis}.
Finally, Section \ref{app:supp_explain_modules} provides a more detailed analysis of the frequency allocation and further elaborates on Figure \ref{fig:period_mono}.

\section{MORE EXPERIMENTS} \label{app:MORE_EXPERIMENTS}
\subsection{Supplementary Ablation Experiments}\label{app:ablation_study}
\noindent \textbf{Ablation Studies on the Scaling Factor $\boldsymbol{\delta}$ for ATS.} \label{app:ATS}
We conduct a series of experiments to further investigate the impact of the temporal scaling factor $\delta$ on the alignment between video and text representations. Accurate temporal alignment plays a vital role in enhancing the model’s understanding of both semantic and sequential aspects of video-language data. To this end, we evaluate the model's performance across three representative video-language benchmarks—\textbf{LongVideoBench}, \textbf{MLVU}, and \textbf{VideoMME}—by varying the temporal scaling factor $\delta$ from 0.5 to 3.0. As shown in Table \ref{tab:ablation_t_steps_2}, we observe a consistent trend across all benchmarks: performance improves as $\delta$ increases, peaking at $\delta = 2$ with an average score of \textbf{60.92}. These results suggest that setting $\delta = 2$ strikes the best balance between temporal resolution and semantic alignment, resulting in optimal overall performance.
\begin{table}[!h]
\setlength\tabcolsep{6pt}  
\centering
\caption{Performance under different scaling factors $\delta$ across multiple benchmarks.}
\label{tab:ablation_t_steps_2}
\vspace{2mm}
\normalsize
\begin{tabular}{ccccc}
\hline
\textbf{Scaling Factor $\boldsymbol{\delta}$} & \textbf{LongVideoBench} & \textbf{MLVU} & \textbf{VideoMME} & \textbf{Avg} \\
\hline
0.5 & 50.83 & 59.87 & 58.33 & 56.34 \\
1.0 & 54.11 & 63.54 & 59.67 & 59.11 \\
\rowcolor[HTML]{F2F3F5}
2.0 & 55.50 & 65.59 & 61.67 & \textbf{60.92} \\
3.0 & 53.83 & 63.38 & 60.33 & 59.18 \\
\hline
\end{tabular}
\end{table}


\paragraph{Ablation Studies on \(\mathbf{x}\), \(\mathbf{y}\) Allocation.} \label{app:x_y_allocation}
To further investigate the impact of different allocation strategies, we conduct quantitative experiments on our proposed \textbf{\methodname}, comparing sequential and interleaved allocations of $x$ and $y$. The results, summarized in Table \ref{tab:ablation_x_y_allocation}, indicate that interleaving $x$ and $y$ leads to superior performance. 

We hypothesize that this improvement arises because interleaving maintains the similarity between the $x$ and $y$ dimensions, whereas sequential allocation increases their disparity, thereby hindering model performance.  
\renewcommand{\arraystretch}{1.1}
\begin{table*}[!h]
\setlength\tabcolsep{5pt}
\centering
\captionsetup{width=0.9\textwidth}
\caption{Ablation Study on \(x\), \(y\) Allocation. \textbf{VideoRoPE (Sequential)} represents the sequential allocation of \(x\) and \(y\), following the pattern \(x, x, x, \dots, y, y, y, \dots\) (similar to M-RoPE~\cite{wang2024qwen2}). \textbf{VideoRoPE (Interleaved)} represents the interleaved allocation, following the pattern \(x, y, x, y, \dots\) (similar to \citet{agrawal2024pixtral12b}).}
\label{tab:ablation_x_y_allocation}
\vspace{2mm}
\normalsize
\begin{tabular}{clllllllll}
\hline
\multirow{2}{*}{\textbf{Method}}  & \multicolumn{4}{c}{\textbf{LongVideoBench}} & \multicolumn{4}{c}{\textbf{MLVU}} \\ 
\cmidrule(lr){2-5} 
\cmidrule(lr){6-9} 
 & 8k & 16k & 32k & 64k & 8k & 16k & 32k & 64k \\ \hline
VideoRoPE(Sequential) & 53.73 & 53.52 & 54.97 & 54.77 & 62.75 & 63.31 & 62.75 & 63.08  \\
\rowcolor[HTML]{F2F3F5} 
VideoRoPE (Interleaved) & {\textbf{54.46}} & \textbf{55.29} & {\textbf{57.15}} & {\textbf{57.26}} & {\textbf{65.19}} & \textbf{66.29} & {\textbf{66.02}} & {\textbf{65.56}} \\ \hline
\end{tabular}

\end{table*}


\paragraph{Ablation Studies on Diagonal Layout Validated on More Benchmarks} \label{app:diagnal_layout}
To further substantiate our claim that the Diagonal Layout (DL) enhances a model’s capability in video understanding tasks, we conduct additional ablation studies on four diverse and challenging benchmarks: \textbf{MLVU}, \textbf{VideoHallucer}, \textbf{V-NIAH}, and \textbf{V-NIAH-D}. These benchmarks cover a wide range of evaluation perspectives, from multi-level video question answering to hallucination detection and fine-grained temporal alignment. As shown in Table~\ref{app:diagonal_layout}, incorporating the DL module consistently improves performance over the baseline model across all benchmarks. Specifically, we observe notable gains on MLVU and V-NIAH and V-NIAH-D, suggesting that DL effectively facilitates better temporal reasoning and semantic alignment. These results reinforce the generalizability and robustness of the proposed Diagonal Layout design in understanding across various tasks.
\begin{table}[!h]
\setlength\tabcolsep{8pt} 
\centering
\caption{Effect of Diagonal Layout (DL) across multiple benchmarks.}
\label{app:diagonal_layout}
\vspace{2mm}
\begin{tabular}{lcccc}
\hline
\textbf{Method} & \textbf{MLVU} & \textbf{VideoHallucer} & \textbf{V-NIAH} & \textbf{V-NIAH-D} \\
\hline
baseline & 61.56 & 34.3 & 78.67 & 74.67 \\
\rowcolor[HTML]{F2F3F5}
+ DL     & \textbf{63.03} & \textbf{34.8} & \textbf{80.44} & \textbf{76.44} \\
\hline
\end{tabular}
\end{table}

\paragraph{Ablation Studies on Different Frequency Allocation Strategies}
\label{app:frequency_allocation}
We compare three different frequency allocation strategies: the \textit{M-RoPE} approach, which emphasizes high-frequency modeling of temporal information and follows a \texttt{[t\ldots{}x\ldots{}y\ldots{}]} format; an interleaved and evenly distributed pattern such as \texttt{[t t x y x y x y]}; and our proposed \textit{VideoRoPE} method, which prioritizes positional encoding followed by low-frequency temporal modeling, arranged in a \texttt{[x y\ldots{}t\ldots{}]} format.

We evaluate these approaches on the \textit{LongVideoBench} benchmark under varying context lengths. This benchmark includes a diverse set of video scenarios, ranging from rapidly changing dynamic scenes to slowly evolving static content.

As shown in the results below, our low-frequency temporal allocation consistently outperforms the interleaved \texttt{[t t x y x y x y]} pattern on average. This suggests that our frequency design more effectively balances global temporal context modeling with local spatial dynamics, making it better suited to handle a wide variety of video conditions.
\begin{table}[!h]
\setlength\tabcolsep{10pt}        
\centering
\caption{Comparison of different frequency allocation strategies under various context lengths.}
\label{app:frequence_allocation}
\vspace{2mm}
\begin{tabular}{lccc}
\hline
\textbf{Context} & \textbf{[t...x...y...]} & \textbf{[t t x y x y]} & \cellcolor[HTML]{F2F3F5}\textbf{[xy...t...] (Ours)} \\
\hline
16k & 60.05 & 59.95 & \cellcolor[HTML]{F2F3F5}\textbf{62.03} \\
32k & 59.33 & 58.40 & \cellcolor[HTML]{F2F3F5}\textbf{59.54} \\
64k & 58.71 & 57.73 & \cellcolor[HTML]{F2F3F5}\textbf{59.12} \\
\textbf{Avg} & 59.36 & 59.06 & \cellcolor[HTML]{F2F3F5}\textbf{60.14} \\
\hline
\end{tabular}
\end{table}

\subsection{Extrapolation to 128k Experiments}
To explore the extrapolation limits of our approach, we extend the visual context during inference to 128k. Specifically, we utilize the \textbf{vLLM framework}~\cite{kwon2023efficient} in Server-API processing mode to enable efficient 128k inference.  

Due to the prolonged evaluation time required for 128k processing, we focus on the \textbf{LongVideoBench} benchmark. As shown in Table \ref{tab:128k-appendix}, although all four methods exhibit performance degradation at 128k, our proposed \textbf{\methodname} experiences the least drop, demonstrating its robustness under extreme extrapolation settings.  

\renewcommand{\arraystretch}{1.1}  
\begin{table}[!h]
\setlength\tabcolsep{5pt}  
\centering
\normalsize
\caption{Comparison of model performance at 64k and 128k context lengths for different methods.}
\label{tab:128k-appendix}
\vspace{2mm}
\begin{tabular}{lcc}
\hline
\multirow{2}{*}{\textbf{Method}} & \multicolumn{2}{c}{\textbf{LongVideoBench}} \\ 
\cmidrule(lr){2-3} 
 & 64k & 128k \\ \hline
Vanilla RoPE~\cite{su2024roformer} & 54.04 & 48.01 \\
TAD-RoPE~\cite{gao2024tc} & 53.42 & 45.77 \\
M-RoPE~\cite{wang2024qwen2} & \underline{54.35} & \underline{51.45} \\
\hline
\rowcolor[HTML]{F2F3F5}  
VideoRoPE & \textbf{57.26} & \textbf{55.64} \\ \bottomrule
\end{tabular}
\end{table}


\section{Additional Details on Evaluation Benchmarks} \label{appendix:benchmarks}

For \textbf{long video understanding}, we employ three benchmarks: (1) \textbf{LongVideoBench} highlights reasoning questions that depend on long frame sequences, which cannot be effectively addressed by a single frame or a few sparse frames, with durations ranging from 8 seconds to 1 hour. We retain only the questions that are free from subtitles.
(2) \textbf{MLVU} provides a comprehensive benchmark tailored for assessing the performance of Multimodal Large Language Models in understanding long videos. The dataset features videos lasting between 3 minutes and 2 hours, with nine diverse evaluation tasks. For our analysis, we concentrate on seven multiple-choice tasks, including Topic Reasoning, Anomaly Recognition, Needle QA, Ego Reasoning, Plot QA, Action Order, and Action Count.
(3) \textbf{Video-MME} stands out as a high-quality benchmark curated for broad scenario coverage, with videos drawn from six key visual domains and 30 subfields. Its dataset spans a wide temporal range, including short clips of 11 seconds and extended videos lasting up to 1 hour.

For \textbf{long video retrieval}, we adopt the following two benchmarks: (1) \textbf{V-NIAH} is specifically designed to identify highly specific moments within long videos, simulating real-world scenarios where only a small segment of a video is relevant within a vast corpus. The setup follows the same configuration as LongVA, where a ``needle'' image is inserted at a random position within a ``haystack'' of 3,000 frames. Each needle image corresponds to a particular question, which is unrelated to the content of the haystack. Each frame is encoded with 144 tokens, and the needle frame is inserted at 0.2 depth intervals. Validation begins at 100 frames, with checks every 200 frames up to 3,000. (2) \textbf{Vision Needle-in-a-Haystack with Distractors (V-NIAH-D)}, our proposed method, builds upon V-NIAH by periodically inserting a distractor 200 frames away from the needle. This distractor is semantically similar to the needle, but it remains irrelevant to the specific question being asked. The insertion period for the distractor is calculated using \( 2 \cdot \pi \cdot 1000000^{32/128} \approx 198.7 \). In our experiments, we directly use a period of 200 for distractor insertion. For additional examples, refer to Figure \ref{fig:v_niah_d_examples}.

For the \textbf{video hallucination}, we use \textbf{VideoHallucer} for evaluation. VideoHallucer classifies hallucinations into two primary types: intrinsic and extrinsic. It further breaks these down into subcategories for detailed analysis, including object-relation, temporal, semantic detail, extrinsic factual, and extrinsic non-factual hallucinations. This framework assesses the model’s ability to accurately answer both basic and hallucinated questions about the video content.


\section{More Related Works}\label{app:related_work}
\noindent \textbf{Related Work on Video LLMs (Video Large Language Models)}
Video Large Language Models (Video LLMs) build upon the success of image-based vision-language models (VLMs)~\cite{liu2023llava,internlmxcomposer,internlmxcomposer2,internlmxcomposer2_5,internlmxcomposer2_5_reward,chen2023sharegpt4v,chen2024open,liu2024rar,liu2024mmdu,huang2024operaalleviatinghallucinationmultimodal,liu2024mia,xing2024pyramiddropacceleratinglargevisionlanguage,zhao2025omnialignvenhancedalignmentmllms,ding2025mm}, which align vision and language representations~\cite{radford2021learningtransferablevisualmodels,zhang2024longclip,sun2023alphaclip} but primarily focus on static images. Extending these models to video requires handling temporal dependencies~\cite{xu2021videoclipcontrastivepretrainingzeroshot,lei2021moreclipbertvideoandlanguagelearning,bertasius2021spacetimeattentionneedvideo,huang2023vtimellmempowerllmgrasp} and long-form video understanding~\cite{wang2024longllavascalingmultimodalllms,chen2024longvilascalinglongcontextvisual,zhang2024longva}. Early video LLMs, such as \citet{wang2023chatvideotrackletcentricmultimodalversatile} and \citet{2023videochat}, leverage various Video Foundation Models (ViFMs), such as InternVideo~\cite{wang2022internvideogeneralvideofoundation}, to extract video attributes, enabling LLM-based question answering. However, their ability to process video content is constrained by the limitations of ViFMs, restricting their effectiveness to short videos. To address this, \citet{luo2023valleyvideoassistantlarge} introduces a Temporal Modeling Module, allowing end-to-end training of LLMs on video data. Building on this approach, \citet{Maaz2023VideoChatGPT} further enhances spatiotemporal modeling to improve video comprehension. Meanwhile, \citet{zhang2023videollamainstructiontunedaudiovisuallanguage} and \citet{lin2023videollava} integrate multiple modalities, such as audio and images, to enrich video understanding. These advancements lay the groundwork for processing long videos with greater accuracy. To extend Video LLMs’ capabilities to longer content, \citet{lin2023mmvidadvancingvideounderstanding} first generates clip-level captions and then employs an LLM to integrate them into a comprehensive video caption, effectively representing the entire video. Various studies, such as those by \citet{li2024llamavid}, \citet{jin2023chatunivi}, \citet{xu2024pllavaparameterfreellava}, and \citet{zhang2025llavaminiefficientimagevideo}, explore different pooling strategies to reduce the number of video tokens, enabling LLMs to process longer videos more effectively. As the field progresses, there is a growing emphasis on long-form video understanding, exploring techniques such as streaming-based processing~\cite{qian2025dispiderenablingvideollms,li2025ovobenchfarvideollmsrealworld}, memory-augmented models~\cite{qian2024streaminglongvideounderstanding,ding2024sam2longenhancingsam2}, and hierarchical representations~\cite{wang2024videotreeadaptivetreebasedvideo} to efficiently model extended temporal structures for tasks like event-level comprehension~\cite{liu2024etbenchopenendedeventlevel} and video summarization~\cite{chai2024auroracapefficientperformantvideo}.

\noindent \textbf{Related Work on Video Haystack Retrieval}
Originating from the Needle-In-A-Haystack task in Natural Language Processing~\citep{niah,multi_niah}, Video haystack tasks aim to locate specific \textit{needle}, the target information, within vast \textit{haystack}, collections of video or multi-modal content~\cite{zhang2024longva,wang2024needle}. In the video domain, VNBench~\cite{zhao2024videoniah} first introduced a video haystack framework with diverse types of needles, such as subtitles, images, and video clips, specifically designed for retrieval tasks within a three-minute timeframe. V-NIAH~\cite{zhang2024longva} further advanced the field by extending retrieval tasks to long durations of up to one hour, providing tools for comprehensive evaluation. In a broader multi-modal domain, MMNeedle~\cite{wang2024multimodal} modeled the retrieval task as locating the exact coordinates of a sub-image within a larger multi-image haystack, while MM-NIAH~\cite{wang2024needle}  introduced a setting where both haystack and needle could include images or text, emphasizing interleaved retrieval capabilities.


However, video haystack retrieval still lags in difficulty compared with the QA or retrieval task in NLP~\citep{hsieh2024ruler,yuan2024lv}. Few attempts exist to enhance the discriminability of evaluations, such as multi-NIAH~\citep{multi_niah,hsieh2024ruler} or NIAH with distractors~\citep{hsieh2024ruler}. Although VHs~\cite{wu2024visual} introduces distractors into the multi-image haystack setting, where needles and distractors are randomly inserted, VHs still did not consider temporal dependencies between video frames or more structured approaches to task evaluation. Our work builds upon V-NIAH by introducing distractors in a systematic, periodic manner based on rotary bases. 

\section{V-NIAH-D Examples}\label{app:video_niah_d}



\begin{figure*}[ht]
    \centering
    \includegraphics[width=0.85\linewidth]{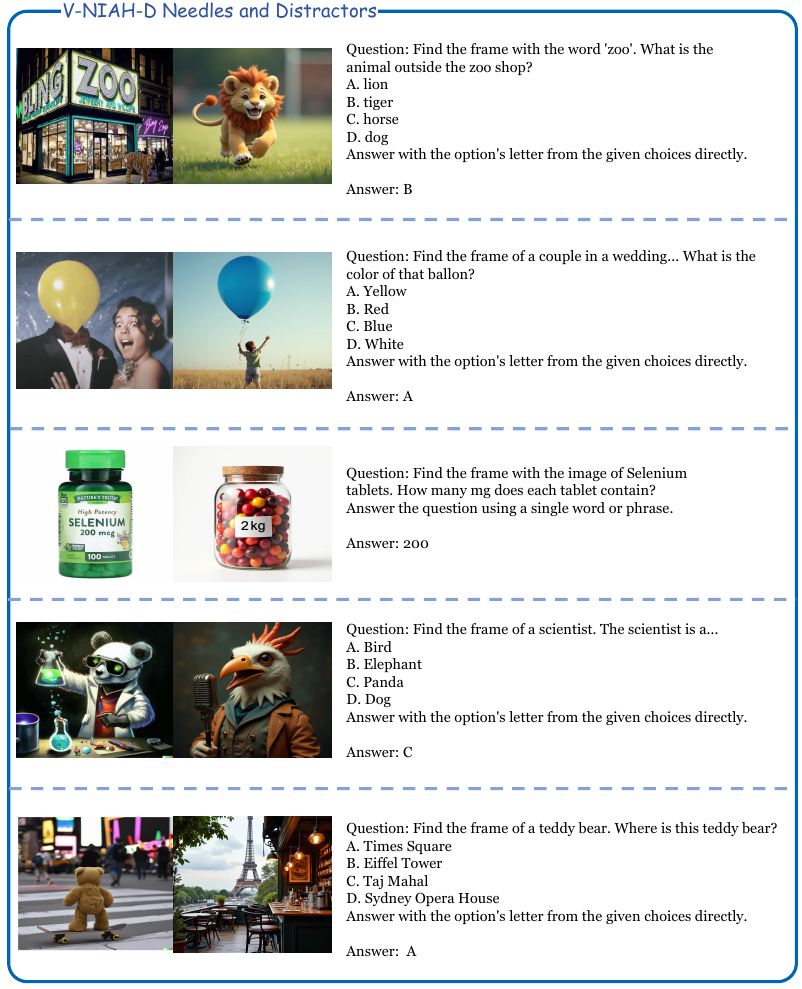}
    \caption{Five visual question-answering problems along with their corresponding needle and distractor.}
    \label{fig:v_niah_d_examples}
\end{figure*}

Figure \ref{fig:v_niah_d_examples} illustrates the five VQA needles used in V-NIAH-D, along with their corresponding distractors. The visual questions and their respective answers are the only components in V-NIAH-D that require human annotation, making it an ideal benchmark for evaluating the long-context reasoning capabilities of LMMs.  

\section{Supplementary Attention Analysis}\label{app:attention_analysis}
\begin{figure*}[!ht]
    \centering
    \includegraphics[width=0.75\linewidth]{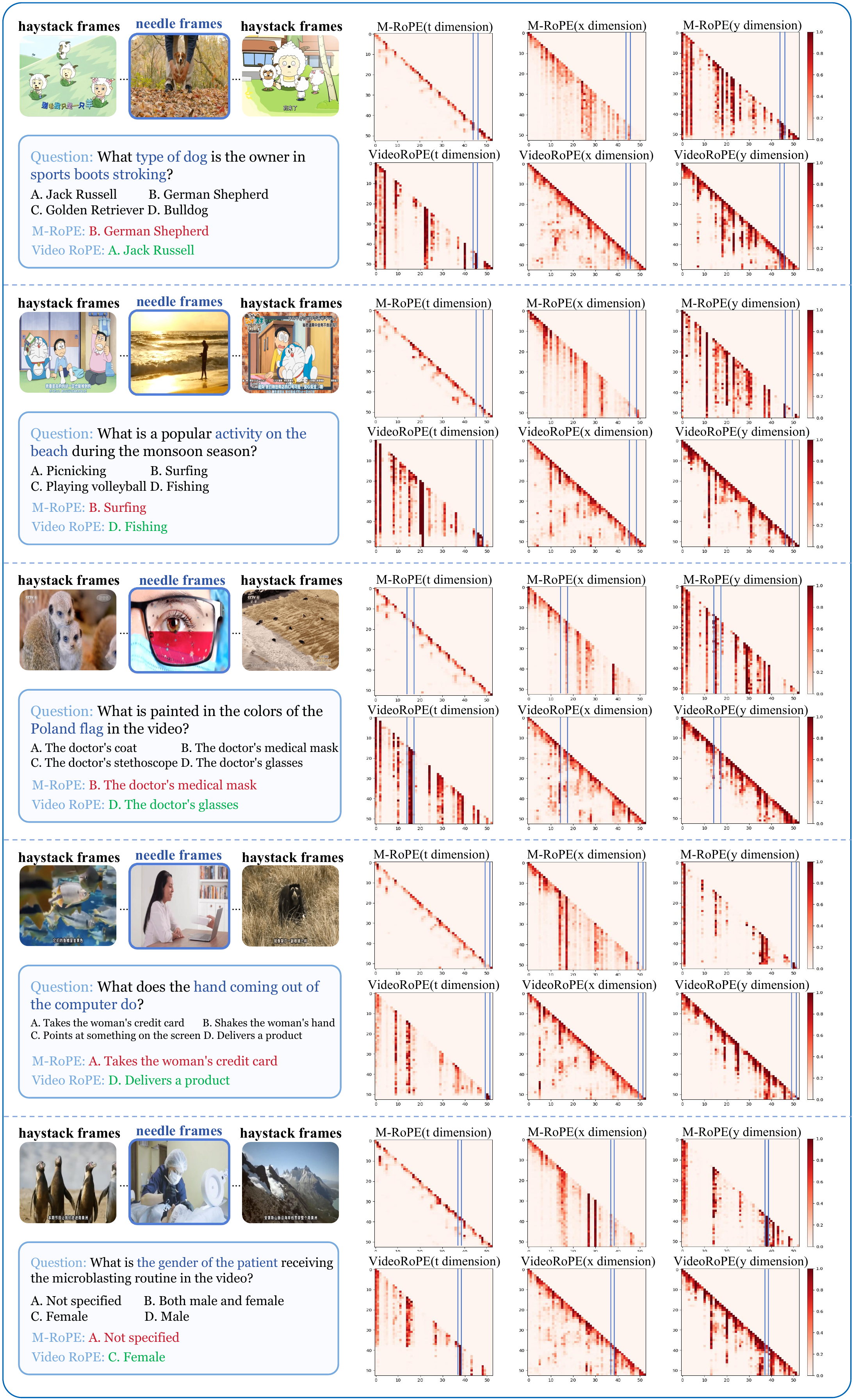}
    \caption{Additional visual analysis of attention.}
    \label{fig:attention_analysis appendix}
\end{figure*}

To further explain the attention pattern in Figure \ref{fig:attention_analysis}, we present additional visual analysis in Figure \ref{fig:attention_analysis appendix}. An attention analysis comparing M-RoPE and \methodname is conducted using 8k-context input, with video tokens from the same frame aggregated through average pooling. As a result, one tick on the axis represents a single frame during inference. The evaluation setup for Figure \ref{fig:attention_analysis} is the same as for Figure \ref{fig:attention_analysis appendix}. M-RoPE relies on high-frequency temporal modeling, limiting it to local information and hindering effective needle identification for question answering. On the other hand, \methodname employs low-frequency temporal modeling, allowing it to capture long-range dependencies and successfully identify the needle for accurate responses.

\section{Supplementary Explanation on Frequency Allocation} \label{app:supp_explain_modules}

This section provides a detailed explanation of the supplementary information related to Figure \ref{fig:period_mono}, highlighting the advantages of our frequency allocation. Consider a RoPE-based LLM with a head dimension size of 128, corresponding to 64 rotary angles $\theta_n$ across different dimensions. In each illustration, we visually represent the function $\cos(\theta_n t)$ for 3 dimensions using parallel blue planes. 

\textbf{(a)} For M-RoPE~\cite{wang2024qwen2}, temporal dependency is modeled using the first 16 rotary angles, which exhibit higher frequency and greater oscillation. Taking the last 3 rotary angles as an example, the position embedding for temporal modeling undergoes significant distortion due to periodic oscillations~\cite{men2024base}, as these dimensions have shorter monotonous intervals. Lower dimensions have even shorter intervals. Notably, because the oscillation is periodic, two distant positions can have nearly identical position embeddings, resembling a hash collision, as shown by the red planes. This phenomenon is why distractors can easily mislead the model.

\textbf{(b)} In contrast, for \methodname, temporal dependency is modeled using the last 16 rotary angles, which have much wider monotonous intervals. Taking the first 3 rotary angles as an example, the position embedding for temporal modeling is free ferom oscillation~\cite{men2024base}. As a result, the misleading effects of distractors are significantly suppressed.

\end{document}